\documentclass[lettersize,journal]{IEEEtran}
\usepackage{newunicodechar}
\newunicodechar{。}{.}

\usepackage{array}
\usepackage{textcomp}
\usepackage{stfloats}
\usepackage{url}
\usepackage{verbatim}
\usepackage{graphicx}
\usepackage{cite}
\usepackage{graphicx}
\usepackage{amssymb}
\usepackage{pgfplots}
\usepackage{cite}
\usepackage{subcaption}
\usepackage{amsthm}
\usepackage{textcomp}
\usepackage{xcolor}
\usepackage{url}
\usepackage{enumerate}

\usepackage[ruled,vlined,boxed]{algorithm2e}
\usepackage{parskip}
\usepackage{tabularray}
\usepackage{arydshln}
\usepackage{wrapfig}
\usepackage{color}
\usepackage{algpseudocode}
\usepackage{mathtools}
\usepackage{multirow}

\newcommand{\kc}[1]{{\color{black}#1}}

\definecolor{DarkBlue}{rgb}{0,0.1,0.7}
\definecolor{DarkRed}{rgb}{0.6,0,0}
\definecolor{DarkGreen}{rgb}{0,0.5,0}

\newtheorem{theorem}{Theorem}[section]

\DeclareMathOperator*{\argmin}{arg\,min}

\newcommand{\tol}{\text{\texttt{tol}}}

\newcommand{\len}{\text{\texttt{len}}}
\newcommand{\inc}{\text{\texttt{inc}}}

\def\BibTeX{{\rm B\kern-.05em{\sc i\kern-.025em b}\kern-.08em
    T\kern-.1667em\lower.7ex\hbox{E}\kern-.125emX}}

\begin{document}

\title{LLM-ABBA: Understanding time series via symbolic approximation}

\author{Xinye Chen, Erin Carson, and~Cheng Kang

\thanks{X. Chen is with Sorbonne Université, CNRS, LIP6, Paris, France, e-mail: xinye.chen@lip6.fr.}

\thanks{E. Carson is with the Department of Numerical Mathematics, Charles University, Prague, Czech Republic, e-mail: carson@karlin.mff.cuni.cz.}

\thanks{C. Kang is with the Department of Cybernetics, Czech Technical University in Prague, Prague, Czech Republic, e-mail: kangchen@fel.cvut.cz.}

\thanks{The first author acknowledges funding from the European Union (ERC, inEXASCALE, 101075632), and additionally acknowledges funding from the Charles University Research Centre program No. UNCE/24/SCI/005.

The second author acknowledges funding from the France 2030 NumPEx Exa-MA (ANR-22-EXNU-0002) project managed by the French National Research Agency (ANR). 

The third author acknowledges funding from the Research Center for Informatics (No.
CZ.02.1.01/0.0/0.0/16\_019/0000765) managed by the Czech Technical University, and additionally acknowledges funding from the Brain Dynamics (No.
CZ.02.01.01/00/22\_008/0004643).

Corresponding author: Cheng Kang.
}
}

\IEEEpubidadjcol
\maketitle

\begin{abstract}
The success of large language models (LLMs) for time series has been demonstrated in previous work. Utilizing a symbolic time series representation, one can efficiently bridge the gap between LLMs and time series. However, the remaining challenge is to exploit the semantic information hidden in time series by using symbols or existing tokens of LLMs, while aligning the embedding space of LLMs according to the hidden information of time series. The symbolic time series approximation (STSA) method called adaptive Brownian bridge-based symbolic aggregation (ABBA) shows outstanding efficacy in preserving salient time series features by modeling time series patterns in terms of amplitude and period while using existing tokens of LLMs. 

In this paper, we introduce a method, called LLM-ABBA, that integrates ABBA into large language models for various downstream time series tasks. By symbolizing time series, LLM-ABBA compares favorably to the recent state-of-the-art (SOTA) in UCR and three medical time series classification tasks. Meanwhile, a fixed-polygonal chain trick in ABBA is introduced to avoid obvious drifting during forecasting tasks by significantly mitigating the effects of cumulative error arising from misused symbols during the transition from symbols to numerical values. In time series regression tasks, LLM-ABBA achieves the new SOTA on Time Series Extrinsic Regression (TSER) benchmarks. LLM-ABBA also shows competitive forecasting capability compared to recent SOTA time series forecasting results. We believe this framework can also seamlessly extend to other time series tasks. Our simulation code is publicly available at:
\begin{center}
    \url{https://github.com/inEXASCALE/llm-abba}.
\end{center}
\end{abstract}


\section{Introduction}
Time series are fundamental mathematical objects with applications across diverse disciplines such as classification \cite{ismail2019deep}, regression \cite{tan2021time}, and prediction \cite{ismail2020benchmarking}. Recently, the power of large language models (LLMs) in time series applications has been recognized. One recent review   \cite{jin2024position} concludes that there are three main LLM-based approaches to learn intricate semantic and knowledge representations from time series to perform various tasks. The first approach is to patch and tokenize numerical signals and related text data, followed by fine-tuning on time series tasks \cite{jin2023time,wang2024timemixer}; the second one is preprocessing time series data to fit LLM input spaces by adding a customized \emph{tokenizer} \cite{gruver2024large}; the last one is to build foundation models from scratch, and this approach aims to create large, scalable models, both generic and domain-specific \cite{rasul2023lag,ekambaram2024ttms}.  

These three techniques each come with their own limitations. Patching and tokenizing time series segments can build the mapping between time series and the latent embedding of LLMs, instead of discrete language tokens. When outputting the numerical value, this method should generate the digit one by one, which eventually reduces the generation speed \cite{jin2023time}. Furthermore, by adding a customized tokenizer, LLMs can handle positions of time series patterns and reproduce the internal logic of given time series signals \cite{mirchandani2023large}. Because LLM tokenizers, not designed for numerical values, separate continuous values and ignore the temporal relationship of time series, this method should convert tokens into flexible continuous values \cite{spathis2024first}. It inevitably requires token transitions from time series feature space to the latent embedding space of LLMs and cannot avoid the risk of semantic loss. Building foundational time series models from scratch can essentially solve these problems. But considering that one should balance the high development costs and their applicability, the challenge of expensive training persists should be tackled \cite{jin2024position}.

\begin{figure}
\begin{center}
\includegraphics[width=0.46\textwidth]{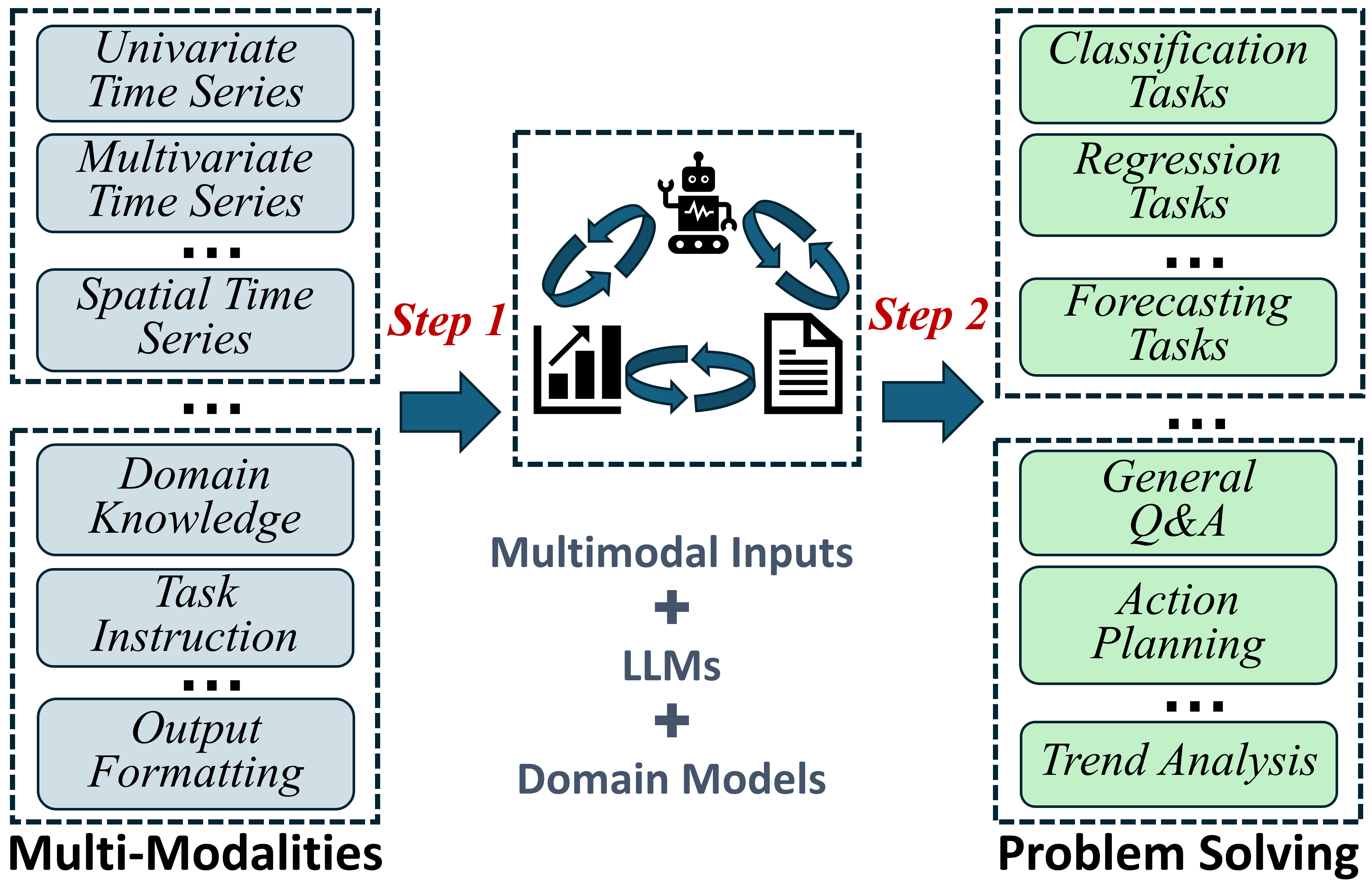}
\end{center}
\caption{The integration of time series and LLM demonstrates potential in solving complex real-world problems. }
\label{fig:LLMtoTS}
\end{figure}

By aligning time series and native language, large language models and specialized time series models constitute a new paradigm, where the LLMs are prompted with both time series and text-based instructions \cite{jin2024position}. Following this paradigm, time series and textual information provide essential contexts, LLMs contribute to internal knowledge and reasoning capabilities, and time series models offer fundamental pattern recognition assurances. This novel integration is depicted in \figurename~\ref{fig:LLMtoTS}, where a successful combination of these components showcases the potential for a general-purpose, unified system in next-generation time series analysis. Therefore, the challenge is to develop one tool that can transform the internal patterns of time series to the contents that LLMs can recognize (\textit{Step 1} of \figurename~\ref{fig:LLMtoTS}). Moreover, this tool should also transform the generated contents back to the time series domain so as to aid the time series analysis (\textit{Step 2} of \figurename~\ref{fig:LLMtoTS}).

Symbolic time series approximation (STSA) is a method that converts time series into symbols. It establishes a bridge between strings and numerical time series, which enables the chain-of-pattern (COP) of strings to be as informative as possible compared to raw data. Utilizing the symbolic representation of time series, one can model time series as native languages by encoding time series as a sequence of strings and performing efficient text analysis techniques upon it rather than manipulating raw numerical values, e.g., converting time series forecasting to next-token prediction in text. STSA could both implicitly and explicitly align the time series features with symbols, which enables the manipulation of natural language processing and learning on time series. If possible, there is no necessity to (1) patch and tokenize time series segments, (2) \kc{add an extra customized tokenizer set}, or (3) build  foundational time series models from scratch. Symbolic representations obtained from transformed numerical time series can potentially reveal the linguistic logic hidden inside time series signals, and this technology roadmap is able to provide LLMs with the ability to understand temporal patterns. Therefore, the time series semantic information can be well exploited in LLMs. Inspired by this idea, it is desirable to obtain a method that can efficiently transform numerical time series into symbols, and fine-tune LLMs on time series analysis tasks (e.g., classification, regression, and forecasting).

\begin{figure}[ht]
\centering
\includegraphics[width=0.5\textwidth]{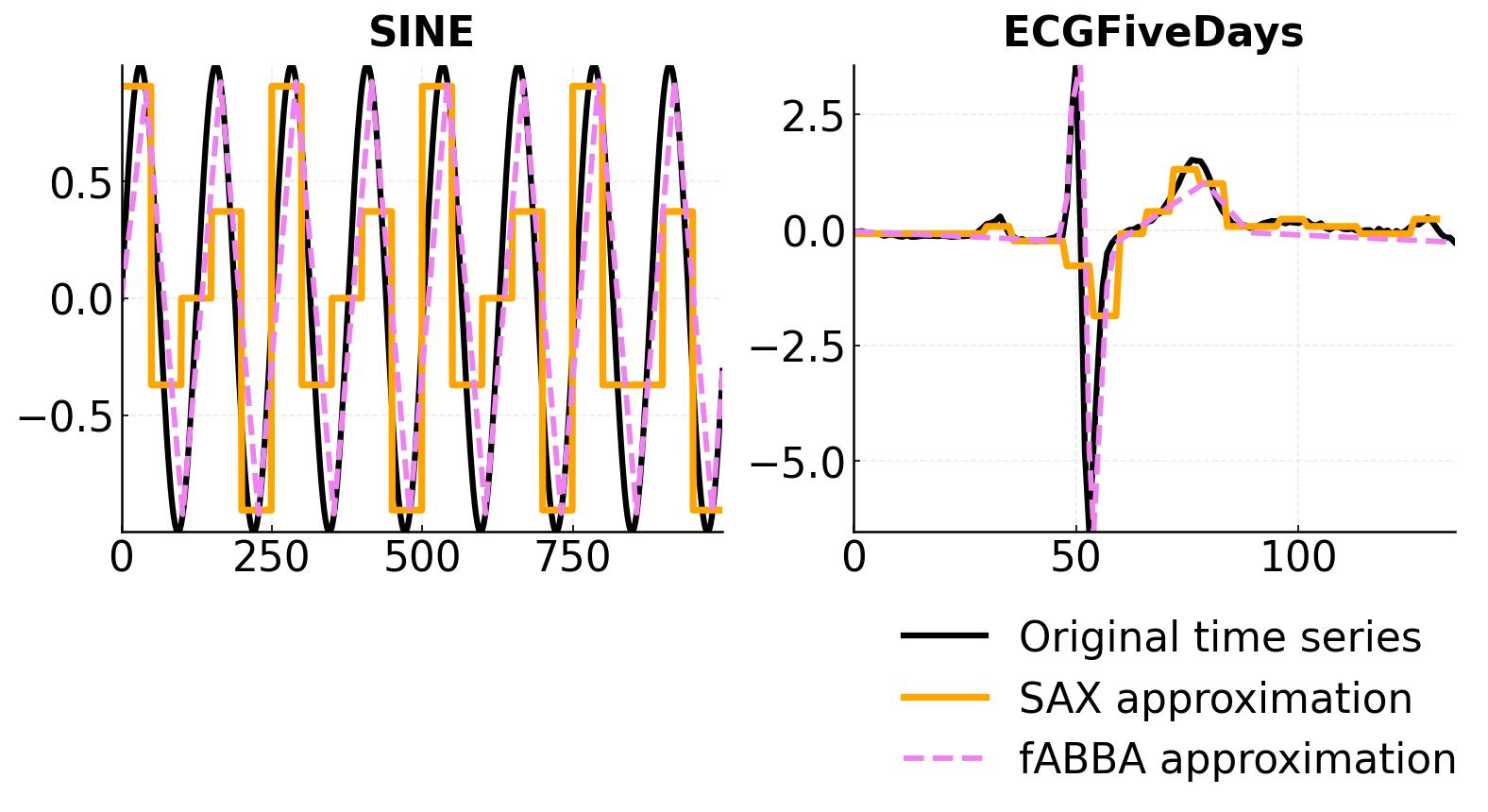}
\caption{The left plot shows a sine function with 1,000 points, and the right plot shows the ECGFiveDays time series from the UCR Archive, which contains 136 points. We first perform fABBA with \tol= 0.1 and $\alpha=0.1$。 Then, we perform SAX with approximately the same length of symbolic representation and the number of distinct symbols to the fABBA. In the sine plot, fABBA generates symbols ``{\color{violet}aBbCbCbCbCbCbCbCA}'' (17 symbols with 5 distinct symbols) while SAX generates symbols ``{\color{violet}aACBbaACBbaACBbaAABb}'' (20 symbols with 5 distinct symbols). In the ECGFiveDays plot, fABBA generates symbols ``{\color{violet}EAbACDBdAcaE}'' (12 symbols with 9 distinct symbols) while SAX generates symbols ``{\color{violet}AAAAAaBBCcADdEaabaaAAb}'' (22 symbols with 9 distinct symbols).}
\label{fig:comp}
\end{figure}

However, the technique to integrate STSA methods with LLMs is lacking. Applying LLMs on symbolic time series representations is tricky. First, we need to address the symbolic consistency issues that exist in STSA methods, as the information of the same symbols across different time series under the same symbolization scheme should be identical. It is also unclear whether LLMs will learn consistent knowledge from the transformed symbols that contain the time series pattern logic. Second, LLMs can generate text contents from given information, but could they also generate symbolic series and reconstruct the time series pattern logic via STSA methods? These considerations bring us to ABBA \cite{EG19b} (incl. its accelerated variant fABBA \cite{fABBA2022}), the most recent STSA method, which shows a competitive advantage in the shape capturing of time series over existing STSA methods. Compared to other STSA methods, ABBA enables users to specify customized strings for symbolization and provides open-source software with easy-to-use APIs\footnote{https://github.com/nla-group/fABBA}. Each ABBA symbol is associated with a unique real-valued cluster center, which enables a natural word embedding for symbols as a native language. A straightforward way to see how much information the STSA methods can capture is via the visualization of their symbolic reconstruction. A comparison of reconstruction using Symbolic Aggregate approXimation (SAX) \cite{lin2007experiencing} and fABBA \cite{fABBA2022} is as illustrated in \figurename~\ref{fig:comp}. It is clear that SAX fails to capture the trend of time series in both figures (also noted in \cite{10.1007/978-3-642-41398-8_24}), and the peak information in figure (b) is missing in the SAX reconstruction. \figurename~\ref{fig:comp} also shows that fABBA is better at capturing the essential information of time series patterns compared to SAX.

In this paper, we propose LLM-ABBA, which can help LLMs to understand time series by using an ABBA method and transforming numerical time series signals into symbolic series. Concretely, LLM-ABBA first transforms time series signals to compressed representations by adaptively compressing numerical inputs. Next, it digitizes the compressed representation with given symbols or pretrained tokens. Then, LLM-ABBA gives LLMs a series of symbols (or pretrained tokens) that LLMs can recognize from the beginning, and these symbols (or pretrained tokens) essentially contain the COPs of time series signals. \kc{Classification tasks only need to identify symbolic series, but for forecasting or regression tasks, an additional step is taken to predict the future time series values.} By using the QLoRA fine-tuning method (a recent, frequently-used adaption fine-tuning method) \cite{dettmers2024qlora}, LLM-ABBA exhibits a trade-off between task performance and efficiency, which also ensures the adaptability on various domains.  Therefore, the LLM is capable of incorporating the COPs of time series and diving into the analysis of time series on a macroscopic view along with the domain knowledge from prompting instructive commands. 
Our contributions include:
\begin{enumerate}
    \item We propose a unified and improved ABBA approach based on fixed-point adaptive piecewise linear continuous approximation (FAPCA) for efficiently symbolizing multiple time series and mitigating the accumulated shift in time series reconstruction, enabling an effective inference task over out-of-sample data.
    \item For time series regression tasks, LLM-ABBA achieves SOTA performance, and it also achieves comparable performance on medical time series classification tasks. To the best of our knowledge, this is the first work to practically combine LLMs with STSA. We believe our work can be easily extended to other STSA methods.
    \item LLM-ABBA can retain language semantics and learn the COPs of time series by adapter fine-tuning methods in time series forecasting tasks. 
    \item The universality and convenience of LLMs' multi-modality on time series tasks obtains a valuable improvement.
\end{enumerate}

The rest of the paper is structured as follows. Section~\ref{sec:related} discusses related work in applications of LLMs to time series. Section~\ref{sec:method} lays the foundation of the ABBA method and proposes our LLM-ABBA framework. Section~\ref{sec:exp} presents the simulations of our method as well as the comparisons between our method and SOTA methods. Section~\ref{sec:limit} discusses the limitations of our method and future work. Section~\ref{sec:conclude} concludes the paper.

\section{Related work}\label{sec:related}

How to bridge the gap between time series and native language is a novel topic. There are existing studies on improving downstream tasks on ABBA symbolic representation rather than raw time series; \cite[Sec.9]{Chen2024thesis} discusses two symbolic forecasting approaches based on ABBA, namely neural network based-approaches and n-gram language models, and gives a straightforward comparison of multiple-step time series forecasting with a feedforward neural network based on ABBA and raw time series. As studied in \cite{EG20b}, the ABBA forecasting framework offers several advantages over directly applying LSTMs to raw time series data: (i) the method effectively compresses the series, removes noise, and captures the essential structural patterns. (ii) This preprocessing step reduces the dimensionality of the input, leading to faster and more stable LSTM training, while also improving robustness to noise and generalization across different time series. Moreover, (iii) the symbolic representation enhances interpretability, as patterns in the symbolic domain can be more easily analyzed and compared than raw numerical signals. Overall, ABBA forecasting yields a more efficient, interpretable, and noise-resistant approach to time series prediction than standard LSTM forecasting on unprocessed data.

LLMs for time series methods have seen significant advances in recent years. The work \cite{gruver2024large} argues that this success stems from the
ability of LLMs to naturally represent multimodal distributions of time series. By framing a time series forecasting task as a sentence-to-sentence task, AutoTimes \cite{liu2024autotimes} minimizes the tunable parameters needed to generate time series embeddings while freezing the parameters of the LLM, and FPT \cite{zhou2023one} fine-tunes LLM parameters to serve as a general representation extractor for various time series analysis tasks. These approaches maximize the use of inherent token transitions, leading to improved model efficiency. In terms of multivariate time series forecasting, 
UniTime \cite{liu2024unitime} trains and fine-tunes a language model to provide a unified forecasting framework across multiple time series domains. Leveraging advanced prompting designs and techniques, PromptCast \cite{xue2023promptcast} transforms time series data into text pairs, and TEMPO \cite{cao2023tempo} models specific time series patterns, such as trends and seasonality, by using weighted scatterplot smoothing \cite{cleveland1990stl}.

Tuning-based predictors use accessible LLM parameters, typically involving pre-processing and tokenizing numerical signals and related prompt text, followed by fine-tuning on time series tasks \cite{jin2024position}. 
In summary, there are four steps needed to adapt LLM to time series: 
\begin{enumerate}[(i)]
    \item $\mathcal{T}_{\mathrm{inp}}=\operatorname{Pre-processing}(\mathcal{T})$: With a Patching operation \cite{liu2024autotimes} or a weighted scatterplot smoothing processing \cite{cao2023tempo}, the time series set $\mathcal{T}$ is pre-processed to specific knowledge-contained inputs $\mathcal{T}_{\mathrm{inp}}$; 
    \item $\mathcal{M}_{\mathrm{inp}}=\operatorname{Tokenizer}(\texttt{Prompt}, \mathcal{T}_{\mathrm{inp}})$: An additional option is to perform a Tokenizer operation on time series $\mathcal{T}_{\mathrm{inp}}$ and related prompt text to form text sequence tokens $\mathcal{M}_{\mathrm{inp}}$;
    \item $\mathcal{M}_{\mathrm{outp}}=f_{\mathrm{LLM}}^{\Delta}\left(\mathcal{M}_{\mathrm{inp}}\right)$: With the instruction prompt $\texttt{Prompt}$, time series processed tokens and optional text tokens are fed into $f_{\mathrm{LLM}}^{\Delta}(\cdot)$ with partial unfreezing or additional adapter layers. $\mathcal{M}_{\mathrm{outp}}$ can be either a fine-tuned result or a intermediate result; 
    \item $\widehat{Y}=\operatorname{Task}\left(\mathcal{M}_{\mathrm{outp}}\right)$: To generate or output required label $\widehat{Y}$, an extra task operation, denoted as Task(·), is finally introduced to perform different analysis tasks. 
\end{enumerate}

\section{Methodologies}\label{sec:method}

Our research is inspired by the observation that speech signals often contain a plethora of semantic information \cite{vandenoord16_ssw}, which enables the language model to perform extremely well across a multitude of tasks; see \cite{jin2024position} and references therein. However, directly applying language models to time series is not possible due to the fact that time series are made up of numerical values and lack useful embedding patterns; further, the high dimensionality of time series makes it difficult for the sequential and recurrent model to capture the dependencies of time series features. Thus learning an informative symbolic time series representation while having dimensionality reduced is a practical yet challenging problem. ABBA{---}a symbolic approximation method{---}is designed to address this as it compresses the time series to a symbolic presentation in terms of amplitude and period, and each symbol describes the oscillatory behavior of time series during a specific period. %

\subsection{ABBA symbolic approximation}

ABBA utilizes adaptive polygonal chain approximation followed by mean-based clustering to achieve symbolization of time series. The reconstruction error of the representation can be modeled as a \emph{Brownian bridge} with pinned start and end points.  ABBA symbolization contains two dominant procedures, namely \emph{compression} and \emph{digitization}, to aggregate a time series $T = [ t_1, t_2, \ldots, t_n] \in \mathbb{R}^{n}$ into its symbolic representation $A = a_1 a_2 \ldots a_N$ where $N \ll n$ and $a_i$ is an element in a specific letter set $\mathcal{L}$, which is referred to as a \emph{dictionary} in the ABBA procedure.

\subsubsection{Compression}

The ABBA compression is performed to compute an adaptive piecewise linear continuous approximation (APCA) of $T$. The ABBA compression plays a critical role in dimensionality reduction in ABBA symbolic approximation{---}a user-specific tolerance, denoted by $\texttt{tol}$, is given to determine the degree of the reduction. The ABBA compression proceeds by adaptively selecting $N+1$ indices $i_0 = 0 < i_1 <\cdots < i_N = n$ given a tolerance $\texttt{tol}$ such that the time series $T$ is well approximated by a polygonal chain going through the points $(i_j , t_{i_j})$ for $j=0,1,\ldots,N$. This leads to a partition of $T$ into $N$ pieces $p_j=(\len_{j}, \inc_{j})$ that represents cardinality and increment of $T_{i_{j-1}:i_j} = [ t_{i_{j-1}},t_{i_{j-1}+1},\ldots, t_{i_j} ]$, which is calculated by $\texttt{len}_j \in \mathbb{N} := i_j - i_{j-1}\geq 1$ and $\texttt{inc}_j \in \mathbb{R} := t_{i_j} – t_{i_{j-1}}$. As such, each piece $p_j$  is represented by a straight line connecting the endpoint values $t_{i_{j-1}}$ and $t_{i_j}$.  Given an index $i_{j-1}$ and starting with $i_0 = 0$, the procedure seeks the largest possible $i_j$ such that $i_{j-1} < i_j\leq n$ and 
\begin{equation}\label{eq:compress}
\begin{aligned}
    \sum_{i=i_{j-1}}^{i_j} \Big( \, t_{i_{j-1}} + (t_{i_j}& - t_{i_{j-1}})\cdot \frac{i - i_{j-1}}{i_j - i_{j-1}}  - t_i \Big)^2 \\&\leq (i_{j} - i_{j-1} -1)\cdot\texttt{tol}^2.
\end{aligned}
\end{equation}
This means that this partitioning criterion indicates that the squared Euclidean distance of the values in $p_j$ from the straight polygonal line is upper bounded by $(\texttt{len}_j - 1)\cdot\texttt{tol}^2$. 

Following the above, the whole polygonal chain can be recovered exactly from the first value $t_0$ and the tuple sequence $[p_1, p_2, \ldots, p_N]$ in the sense that the reconstruction error of this representation is with pinned start and end points and can be naturally modeled as a Brownian bridge. In terms of \eqref{eq:compress}, a lower  $\texttt{tol}$ value is required to ensure an acceptable compression of time series with a great variety of features such as trends, seasonal and nonseasonal cycles, pulses and steps. As indicated in \cite{EG19b}, the error bound between the reconstruction and original time series is upper bounded by $(n - N)\cdot \tol^2$.

\subsubsection{Digitization}
The ABBA compression is followed by a reasonable digitization that leads to a \emph{symbolic representation}. Prior to digitizing, the tuple lengths and increments are separately normalized by their standard deviations $\sigma_{\texttt{len}}$ and $\sigma_{\texttt{inc}}$, respectively. After that, further scaling is employed by using a parameter $\texttt{scl}$ to assign different weights to the length of each piece $p_i$, which denotes the importance assigned to its length value in relation to its increment value. Hence, the clustering is effectively performed on the \emph{scaled tuples} $p^s_i=\left(\texttt{scl}\frac{\texttt{len}_i}{\sigma_{\texttt{len}}}, \frac{\texttt{inc}_i}{\sigma_{\texttt{inc}}}\right)$, $i=1, \ldots, N$.  In particular, if $\texttt{scl} = 0$, then clustering will be only performed on the increment values of $p^s_i$, while if $\texttt{scl} = 1$,  the lengths and increments are treated with equal importance.

The next step after normalization works with a mean-based clustering technique in Euclidean space. In the ABBA setting, letting the input of $N$ vectors be $P^s=[p^s_1, \ldots, p^s_N] \in \mathbb{R}^{2\times N}$, one seeks a codebook of $k$ vectors, i.e., $C=[c_1, \ldots, c_k] \in \mathbb{R}^{2 \times k}$ ($k \ll N$) where each $c_i$ is associated with a unique cluster $S_i$ such that $k$ clusters from $P^s$ minimize the sum of Euclidean distances $\texttt{SSE}$ constructed by $C$. The obtained codebook vectors are known as cluster centers. A quality codebook produces $k$ clusters $S_1, S_2, \ldots, S_k \subseteq P^s$ such that the sum of squared errors $\texttt{SSE} =  \sum_{i=1}^{k}\sum_{p^s \in S_i}\|p^s - c_i\|_2^2$ is small enough to an optimal level. However, this is a suboptimal solution to minimizing $\texttt{SSE}$. The k-means problem aims to find $k$ clusters within data in $d$-dimensional space, to minimize the \texttt{SSE}. As the iterations proceed, the mean value $\mu_i:= \frac{1}{|S_i|}\sum_{p^s \in S_i} p^s$ is always chosen for updating centers $c_i$ in k-means algorithms\cite{journals/tit/Lloyd82} to ensure \texttt{SSE} decreases. However, attainning the global minimum \texttt{SSE} is NP-hard even if $k$ is restricted to $2$ \cite{10.1145/1374376.1374452} or in the plane \cite{MAHAJAN201213}. Typically, the sub-optimal k-means problem in digitization can also be solved by a sorting-based aggregation \cite{fABBA2022}, which achieves a speedup by orders of magnitude compared to the one using k-means (in practice, $\texttt{k-means++}$ is employed). The principle of the aggregation is to achieve an error-controlled clustering by greedily selecting the \emph{starting points} according to a precomputed sorting. An efficient algorithm to symbolize time series in a large-scale way is desired, and thus sorting-based aggregation for a description) is preferred. The number of symbols generated by sorting-based aggregation is determined by the parameter $\alpha$; see  \cite{fABBA2022} for details. The \texttt{SSE} achieved by fast sorting-based aggregation~\cite{fABBA2022} is upper bounded by $\alpha^2(N - k)$ , and the expected \texttt{SSE} value is $\frac{\alpha^2(N - k)}{2}$. 

In the context of symbolic approximation, we refer to cluster centers as \emph{symbolic centers}, and each symbolic center is associated with an identical symbol. Then each $p^s_i$ is assigned to the closest symbolic center $c^i$ associated with its symbol $c^i = \argmin_{c \in C}(\|p^s - c\|)$. After that, each $p^s_i$ is associated with a unique center, which is assigned as a label. We use a symbol to correspond to the label. The symbols can be represented by text characters, which are not limited to English alphabet letters, e.g., ASCII codes or any of its combinations. As such, the ABBA symbolization can be flexibly adapted to LLMs' pretrained tokens. 

\subsubsection{Inverse symbolization}
The \emph{inverse symbolization} step converts the symbolic representation $A$ back to the reconstructed series $\widehat{T}$, which is key for some value forecasting tasks in time series. The inverse symbolization is followed by a \emph{inverse-digitization} that uses the $k$ representative elements $c_i \in C$ to replace the symbols in $A$ and denormalize them separately, thus resulting in a 2-by-$N$ array $\widetilde{P}${---}an approximation of $P$. Each $\widetilde{p}_i \in \widetilde{P}$ is the closest symbolic center $c^i \in C$ to $p^s_i \in P^s$ (in contrast to $P$) after denormalization. However, the inverse digitization often leads to non-integer values for the reconstructed length $\len$, so a rounding method is used to align the accumulated lengths with the closest integers. The first length is rounded to an integer value, i.e., $\widehat{\len}_1:= \text{round}(\widetilde{\len}_1)$ and the rounding error $e:= \widetilde{\len}_1 - \widehat{\len}_1$ is computed. The error is then added to the rounding of $\widetilde{\len}_2$, i.e., $\widehat{\len}_2 := \text{round}(\widetilde{\len}_2 + e)$, and the new error $e'$ is calculated as $\widehat{\len}_2 + e - \widetilde{\len}_2$. Then $e'$ is similarly involved in the next rounding. After all rounding is computed, we obtain 
\begin{equation}\label{eq:repolygon}
(\widehat{\len}_{1}, \widehat{\inc}_{1}),(\widehat{\len}_{2}, \widehat{\inc}_{2}), \ldots, (\widehat{\len}_{N}, \widehat{\inc}_{N}) \in \mathbb{R}^{2},
\end{equation}
where the increments $\inc$ are unchanged, i.e., $\widehat{\inc} = \widetilde{\inc}$. The last step is to recover $\widehat{P}$ exactly from the initial time value $t_0$ and the tuple sequence \eqref{eq:repolygon}, resulting in the reconstructed time series $\widehat{T}$.

\subsection{Error analysis reconstruction}
We are concerned with the reconstruction error of ABBA's symbolization since a symbolic representation with a higher reconstruction error is a less informative representation. It is worth noting that the reconstruction of time series from the compression procedure proceeds by establishing a polygonal chain $\widetilde{T}$ going through the chosen tuples $\{(i_j, t_{i_j})\}_{j=0}^{N}$ from the original time series $T$ and $\len_j = i_{j+1} - i_{j}$. As indicated in \cite{EG19b}, a polygonal chain $\widehat{T}$ stitching together $\{(\widehat{i}_j, \widehat{t}_{i_j})\}_{j=0}^{N}$ via a tuple sequence $\widehat{P}$ is reconstructed by the inverse symbolization, thus we have Theorem~\ref{thm:reconst}.

\begin{theorem}\label{thm:reconst}
Let $(\mu_i^{\len}, \mu_i^{\inc}) = \frac{1}{|S_i|} \sum_{(\len, \inc) \in S_i} (\len, \inc)$, we denote the mean set for $\len$ and $\inc$ by $\mathcal{U}_{\len}=\{\mu_i^{\len}\}_{i=1}^{k}$ and $\mathcal{U}_{\inc}=\{\mu_i^{\inc}\}_{i=1}^{k}$, respectively. Since $i_0=0$, the reconstruction indices and size of time series values are given by 
\begin{equation}
    (\widehat{i}_j, \widehat{t}_{i_j}) = \bigg(\sum_{\ell=1}^j \widehat{\len}_{\ell}, t_0 + \sum_{\ell=1}^j \widehat{\inc}_{\ell}\bigg), \quad \text{for } j= 0, \ldots, N, 
\end{equation}
where $(\widehat{\len}_{\ell}, \widehat{\inc}_{\ell})$ are the computed cluster centers, i.e., $\widehat{\len}_{\ell} \in \mathcal{U}_{\len}$ and $\widehat{\inc}_{\ell} \in \mathcal{U}_{\inc}$.
\end{theorem}

Theorem~\ref{thm:reconst} shows the accumulated deviations from the true lengths and increments are canceled out (as analyzed in  \cite{EG19b}) at the right endpoint of the last piece $p_N$, thus $(\widehat{i}_N, \widehat{t}_{i_N}) = (i_N, t_{i_N}) = (n, t_n)$, which indicates the start and ending point between $\widehat{T}$, $\widetilde{T}$ and $T$ are identical. We thus have the following result.

We now denote the local deviation of the increment and length by 
\begin{equation}
\begin{aligned}
    d^{\inc}_\ell := \widehat{\inc}_{\ell} - \widetilde{\inc}_{\ell}, \quad  
    d^{\len}_\ell := \widehat{\len}_{\ell} - \widetilde{\len}_{\ell}.
\end{aligned}
\end{equation}
\begin{theorem}[\cite{EG19b}]\label{thm:deviate}
\[\sum_{i} \sum_{(\len, \inc) \in S_i} (d^{\len}, d^{\inc}) = (0, 0).
\]
\end{theorem}

\begin{theorem}\label{thm:deviate2}
Assume that ABBA is performed with hyperparameter $\alpha$, and it results in $k$ clusters $S_1, \ldots, S_k$. Then we have 
\begin{equation}\label{eq:bound_p}
    \max_{\ell}\{(d^{\inc}_{\ell})^2 + (d^{\len}_{\ell})^2\}  \le \alpha^2,
\end{equation}
 and further
\begin{equation*}
\begin{aligned}
\sigma = \max_{i = 1, \ldots, k} \frac{1}{|S_i|}& \sum_{(\len, \inc) \in S_i} \bigg(|\len - \mu_i^{\len}|^2 + \\&|\inc - \mu_i^{\inc}|^2  \bigg) \le \alpha^2,
\end{aligned}
\end{equation*}
\end{theorem}

Following Theorem~\ref{thm:deviate2}, the $\sigma$ is explicitly controlled by $\alpha$, and thus we remove the need to estimate the additional parameter  $\tol_s$ that is used in \cite{EG19b} by directly relating it to the hyperparameter $\alpha$. 

Given the $N$ data points selected by the adaptive polygonal approximation chain, letting $e_{j}^{\len} :=  \sum_{\ell=1}^j d_{\ell}^{\len}$ and $e_{j}^{\inc} :=  \sum_{\ell=1}^j d_{\ell}^{\inc}$, it is obvious that $e_{j}^{\inc} = \widehat{t}_{i_j} - t_{i_j}$ if $e_{j}^{\len}=0$ for $j=1, \ldots, N$. 

\noindent\textbf{Modeling assumption.} 
We assume that the local deviations $\{d^{\len}_\ell, d^{\inc}_\ell\}_{\ell=1}^{N}$ form a sequence of bounded, independent random variables with zero mean, satisfying $(d^{\len}_\ell)^2 + (d^{\inc}_\ell)^2 \le \alpha^2$. 
This independence and boundedness assumption allows the use of Hoeffding-type concentration inequalities and implies that the cumulative deviations $\{e_j^{\len}, e_j^{\inc}\}$ behave as a discrete Brownian-bridge process anchored at $(0,0)$ and $(N,0)$.

This leads to Theorem~\ref{thm:deviate3} and Theorem~\ref{thm:deviate4} below.

\begin{theorem}\label{thm:deviate3}
\begin{equation*}
\begin{aligned}
|e_{j}^{\inc}| \le  j \sqrt{\alpha^2 - (d^{\len}_{\ell})^2} \le   j |\alpha|,
\end{aligned}
\end{equation*}
where $j=0, \ldots, N$. 

Similarly, the shift of the time series has $|e_{j}^{\len}| \le j \sqrt{\alpha^2 - (d^{\inc}_{\ell})^2} \le j |\alpha|$ for $j=0, \ldots, N$.
\end{theorem}

\begin{theorem}\label{thm:deviate4}
For all $h > 0$, 
\begin{equation*}
\mathbb{P}(|e_{j}^{\inc}| \ge  h) \le \exp{(- \frac{h^2}{2j\alpha^2})} 
\end{equation*}\text{and}\begin{equation*}
\mathbb{P}(|e_{j}^{\len}| \ge  h) \le \exp{(- \frac{ h^2}{2j\alpha^2})}.
\end{equation*}

\end{theorem}

\begin{proof}[Proof of Theorem~\ref{thm:deviate4}]

From Theorem~\ref{thm:deviate}, we can easily obtain 
\begin{equation*}
\begin{aligned}
    (e_{0}^{\len}, e_{0}^{\inc}) &= (0,0), \quad (e_{N}^{\len}, e_{N}^{\inc}) &= (0,0)
\end{aligned}
\end{equation*}
associated with expectation $E(e_{j}^{\len}) = E(e_{j}^{\len})  = 0$. 

For $j=1, \ldots, N$, since $ d^{\len}_j, d^{\inc}_j \in [-\alpha, \alpha]$, using \eqref{eq:bound_p} and  Hoeffding's inequality, 
\begin{equation*}
\begin{aligned}
\mathbb{P}(\left|\sum_{\ell=1}^{j} ( d^{\inc}_\ell - E[ d^{\inc}_\ell])\right| \ge h)&=  \mathbb{P}(\left|e_{j}^{\inc} - E[e_{j}^{\inc}]\right| \ge h) \\&\le \exp{(- \frac{ 2h^2}{j\alpha^2})}.
\end{aligned}
\end{equation*}

Therefore, 
$\mathbb{P}(|e_{j}^{\len}| \ge  h) \le \exp{(- \frac{ 2h^2}{ j\alpha^2})}$  \text{and} $\mathbb{P}(|e_{j}^{\inc}| \ge  h) \le \exp{(- \frac{ 2h^2}{j\alpha^2})}$
for all $t > 0$. 
\end{proof}

This means that a decrease of $\alpha$ is prone to result in a smaller reconstruction error $e_{j}$. This phenomenon was mentioned in \cite{EG19b}.  The growth of $j$ increases the possibility of larger errors, since the errors coming from the previous reconstruction will be accumulated into the subsequent reconstruction by the principle of inverse symbolization.

\subsection{ABBA to LLM}
In the following, we write a time series containing $n$ data points as $T_q$, and use $\mathcal{T}=\{T_q\}_{q=1}^{Q}$ to denote a set of time series of dimension $Q$, associated with its corresponding symbolic representation set $\mathcal{A}=\{A_q\}_{q=1}^{Q}$. If $Q=1$, $T$ is an univariate time series.

\begin{figure*}[ht]
\begin{center}
\includegraphics[scale=0.15]{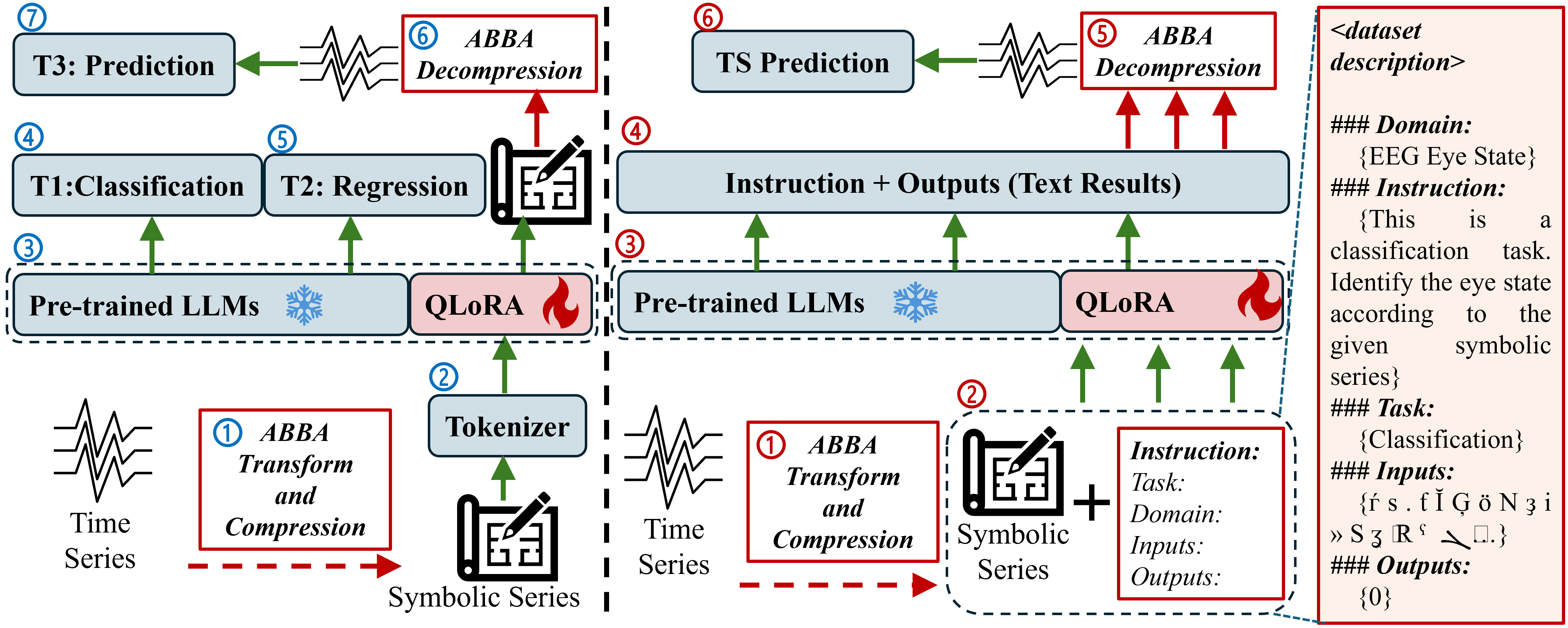}
\end{center}
\caption{The model framework of LLM-ABBA: Given an input time series, we first transform and compress the time series to a symbolic series via \textcolor{blue}{\textcircled{1}} and \textcolor{red}{\textcircled{1}}. These symbolic series will be tokenized by the LLM's tokenizer \textcolor{blue}{\textcircled{2}}. The designed instruction that contains the symbolic series also will be tokenized by the LLM's tokenizer \textcolor{red}{\textcircled{2}}. Additionally, by only fine-tuning the pretrained LLM, the QLoRA with inhibition mechanism is utilized both in \textcolor{blue}{\textcircled{3}} and \textcolor{red}{\textcircled{3}}. To implement the corresponding tasks, \textcolor{blue}{\textcircled{4}} and \textcolor{blue}{\textcircled{5}} loads the LLM according to the type of task. However, \textcolor{red}{\textcircled{4}} loads the LLM on the generation task. Moreover, to inverse symbolic series back to numerical time series, \textcolor{blue}{\textcircled{6}} and \textcolor{red}{\textcircled{5}} utilizes ABBA to decompress the generated symbolic series. Lastly, in \textcolor{blue}{\textcircled{7}} and \textcolor{red}{\textcircled{6}} the output time series from LLM-ABBA are projected to generate the forecasts.}
\label{fig:framework}
\end{figure*}

\subsubsection{Fixed-point adaptive polygonal chain}

\begin{figure}[ht]
\centering 
\subfloat{\includegraphics[width=0.48\textwidth]{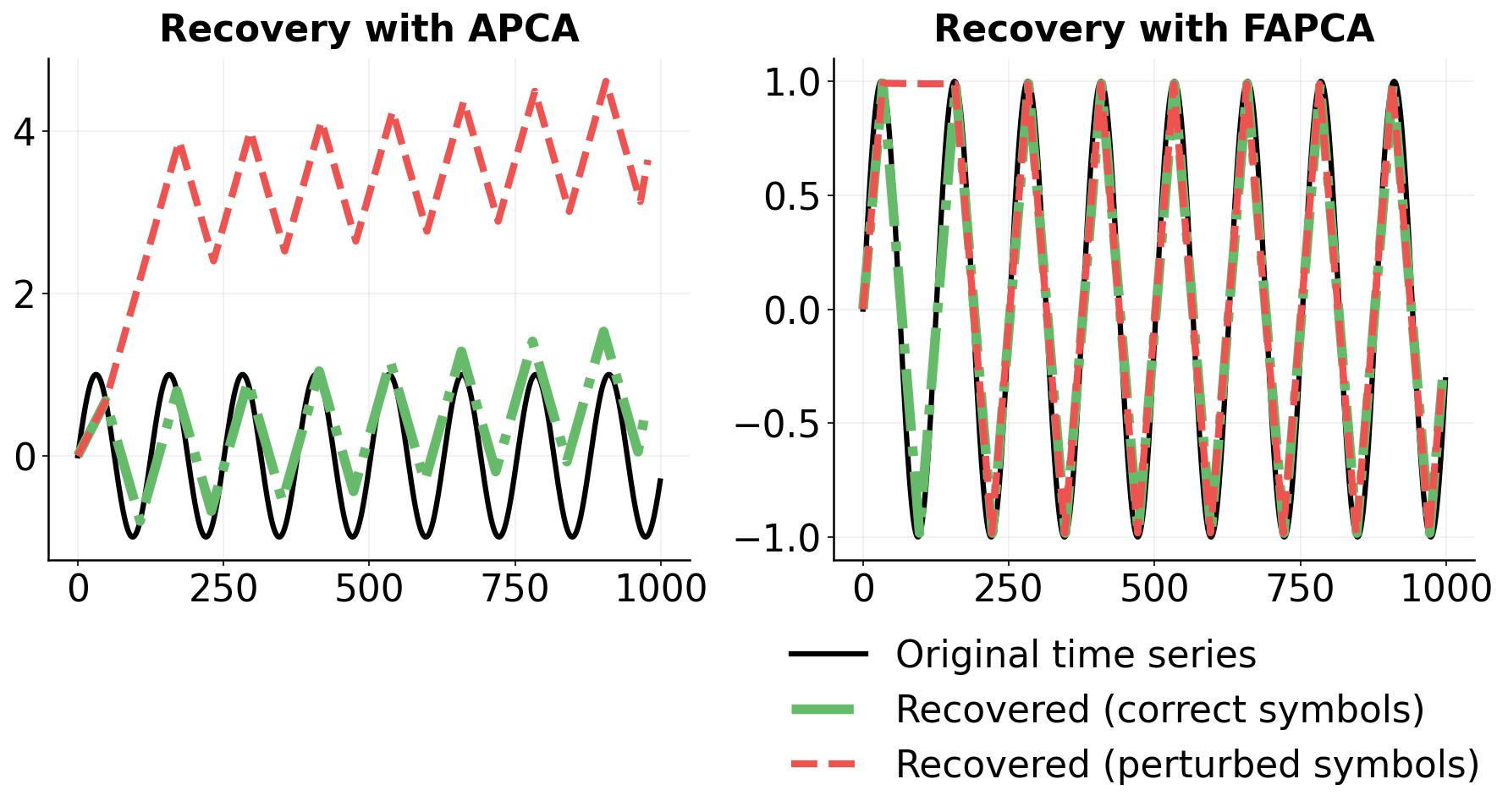}}
\caption{We generate a synthetic trigonometric sine series of 1,000 points, and separately perform symbolic approximation with 4 symbols using APCA (left) and FAPCA (right) on the time series. ABBA with APCA and FAPCA generate symbols ``{\color{violet}aBbBbBbBbBbBbBbBA}'' and ``{\color{violet}abBbBbBbBbBbBbBbA}'', respectively, associated with their respective perturbed symbols, ``{\color{violet}a{\color{red}b}bBbBbBbBbBbBbBA}'' and ``{\color{violet}a{\color{red}B}BbBbBbBbBbBbBbA}''. The symbol recovery is performed on correct symbols and perturbed symbols, respectively.}
\label{fig:newtrick}
\end{figure}

In time series forecasting settings, the value-based forecasting is converted into a token-based forecasting using STSA. However, it is very desirable to mitigate the negative effect of the preceding mistakenly predicted symbol on the subsequent time series recovery since the recovery proceeds from front to back. However, APCA and the symbolic recovery often lead to a cumulative error for symbolic forecasting, that is, \kc{an incorrect replacement} of a previous symbol will influence the subsequent reconstruction. A \emph{fixed-point polygonal chain} trick is introduced to mitigate this issue. We still partition the time series into pieces following \eqref{eq:compress} while $p_j=(\len_{j}, \inc_{j})$ is replaced with  $p_j=(\len_{j}, t_{i_j})$ before normalization. We call the new approximation method \kc{fixed-point adaptive piecewise linear continuous approximation (FAPCA)}. The resulting tuples $p_i$ will be normalized and one can be recovered from the other since $\texttt{inc}_j = t_{i_j} - t_{i_{j-1}}$. \figurename~\ref{fig:newtrick} shows that FAPCA eliminates the cumulative errors arising from the preceding mistaken symbol and improves the recovery.

\subsubsection{Symbolizing multiple/multi-dimensional time series}

SMTS \cite{baydogan2015learning} trained a tree learner to consider all attributes of multiple/multi-dimensional time series, both abbreviated as MTS, as the interactions in the space of the time index and time values can be detected from a generated high-dimensional codebook from the terminal nodes of the trees. Traditional ABBA focuses on converting a univariate time series; it can not convert MTS with consistent symbolic information (i.e., each symbol is associated with a unique symbolic center). Thus, by keeping consistent symbolic information across the dimensions of MTS, \cite{chen2024joint} proposes a joint approach to process MTS efficiently;  in detail, the ABBA approach starts with concatenating the compressed pieces from separately applied adaptive piecewise polygonal chain approximation to each channel/sample and performs digitization on the merged pieces all at once, and allocates the symbol back to each channel/ sample. Based on that, the distinct symbol in each channel will natively have an implicit interaction among MTS and thus each channel/sample, which enables LLM to automatically learn it.

We illustrate a unified approach towards a consistent symbolic approximation for multiple univariate time series:
\begin{itemize}
    \item Step 1: Use APCA or FAPCA to compress each time series $T_q$ into $P_q$ for $q=1,\ldots Q$
    \item Step 2: Compute normalized $P^s_q$ and concatenate $P^s_q$ to form $\mathcal{P}^s := [P^s_q]_{q=1}^Q$
    \item Step 3: Perform digitization 
 on $\mathcal{P}^s$
    \item Step 4: Allocate symbols to each time series (the number of symbols for $T_q$ is equal to $|P^s_q|$)
\end{itemize}

\subsubsection{Symbolizing out-of-sample data}

Symbolizing out-of-sample time series data with consistent symbols is essential for various time series downstream tasks, e.g., inference tasks. To symbolize $\mathcal{T}^{\text{out}}=\{T^{\text{out}}_q\}_{q=1}^{Q'}$, we perform the following steps:
\begin{itemize}
    \item Step 1: Compress each time series $T^{\text{out}}_i$ into $P^{\text{out}}_q$ for $q=1,\ldots Q'$
    \item Step 2: Assign a symbol to each $p \in P^{\text{out}}_q$ following the rule of digitization
\end{itemize}

For multivariate time series, ABBA processes them similarly to multiple univariate time series. The only additional step is to flatten each multivariate series by concatenating its values from all channels (dimensions) in sequence, creating a single one-dimensional sequence. This conversion transforms multiple multivariate time series into several univariate time series. Symbolic sequences are then generated using the same approach as for multiple univariate time series, and the symbols are mapped back to each channel (dimension).

\subsubsection{Feeding the LLM}
ABBA can transform numerical time series to symbolic series and keep the internal logic chain from which LLMs can learn  temporal knowledge. In other words, by ensuring the precondition that the input symbolic series inherits the polygonal chain of numerical time series and then represents this chain via symbolic series (or LLMs' tokens) that can be recognized by LLMs, LLMs can reconstruct the embedding space without the use of any new tokens via adapting fine-tuning methods.

For the connection to LLMs, \figurename~\ref{fig:framework} shows two proposed frameworks, and only the left symmetric one is used. First, the symbolic approximation method ABBA is used to transform the time series into a symbolic series. Second, the tokenizer of large language models is utilized to tokenize the symbolic series. The quantized low-rank adaptation fine-tuning method with the shunting inhibition mechanism is applied to fine-tune the LLMs. In the classification tasks, the LLMs are loaded as a classification-following model, and the cross-entropy loss function is used. In the regression task, the bottom layer of the pre-trained LLMs is replaced by a regression layer, and the RMSE is used as the loss function to fine-tune the LLMs. In forecasting tasks, the LLMs are loaded as a generation-following model. The generated symbols will be transformed back to a numerical series by using the inverse symbolization process of ABBA. The right panel semantics is an extension application. LLMs are trained as a generation-following model, and the additional work involves incorporating more information into the instructions, such as the instruction command, task type, and domain. This instruction design is almost equal to the left panel (without LLMs' Instructions) of \figurename~\ref{fig:framework}.

For the consistency of related tuning-based methods, $\mathcal{T}$ is referred to as the input in the time series dataset, $\mathcal{A}$ is the symbolic representation generated by ABBA; $\phi: \mathcal{T} \rightarrow \mathcal{A}$ denotes the symbolization of ABBA, and $\phi^{-1}: \mathcal{A} \rightarrow \mathcal{T}$ is referred to as the inverse symbolization of ABBA. We formulate the framework of LLM-ABBA:
\begin{enumerate}[(i)]
    \item $\mathcal{A} = \phi(\mathcal{T})$: The input $\mathcal{T}$ is converted to its symbolic representation $\mathcal{A}$. 
    \item $\mathcal{M}_{\mathrm{inp}}=\operatorname{Tokenizer}( \texttt{Prompt}, \mathcal{A})$: Tokenizing the symbolic representation $\mathcal{A}$; here, the $\operatorname{Tokenizer}$ is the default Tokenizer for LLMs. 
    \item $\mathcal{M}_{\mathrm{outp}}=f_{\mathrm{LLM}}^{\Delta}\left(\mathcal{M}_{\mathrm{inp}}\right)$: Feed the tokenized input to LLM model. 
    \item $\widehat{Y}=\operatorname{Task}\left(\mathcal{M}_{\mathrm{outp}}\right)$: If this is a classification task, $\widehat{Y}$ is a generated label. If the task is a regression or forecasting task, $\widehat{Y}$ is an ABBA-transformed numerical value or sequence produced by the inverse symbolization process of ABBA:

\begin{center}
$\begin{cases}
 \widehat{Y}=\mathcal{M}_{\mathrm{outp}}, & \text{Classification task} , \\
\widehat{Y}=\phi^{-1}\left(\mathcal{M}_{\mathrm{outp}}\right), & \text {Regression / forecasting task} 
\end{cases}$
\end{center}

\end{enumerate}

The complexity of ABBA, as discussed in \cite{EG19b}, is dominated by compression and digitization, where the compression takes linear complexity for time series of length $n$ and the digitization runtime is independent of $n$. Rather, for the compressed pieces of cardinality $k$, it uses fast aggregation which takes near-linear complexity $\mathcal{O}(k\log k + kd)$ on average and $\mathcal{O}(k^2 d)$ in the worst case; luckily in our setting, $d=2$ is a constant since our digitization only operates on $\texttt{inc}$ and $\texttt{len}$. In practice we employ a high compression rate so that $k \ll n$. To speed up the symbolization, we apply parallelism for compression which significantly decreases the runtime of the symbolization process.  The training complexity of fine-tuning LLMs was described in LoRA \cite{hu2022lora}. The fine-tuning complexity of LoRA is $\mathcal{O}(n)$, but the inference complexity is $ \mathcal{O}(n^2)$. Here, $n$ is the length of the input sequence of LLMs (or the length of the pieces).

Compared to SAX's complexity of $\mathcal{O}(n)$ and SFA's complexity of $\mathcal{O}(n \log n)$, ABBA typically requires an additional internal data management overhead for adaptive segmentation, but it often yields superior shape preservation which benefits downstream classifiers and LLM integration.

\subsection{Linguistics investigation: Zipf's law}

\begin{figure}[htbp]
\centering
\includegraphics[width=0.47\textwidth]{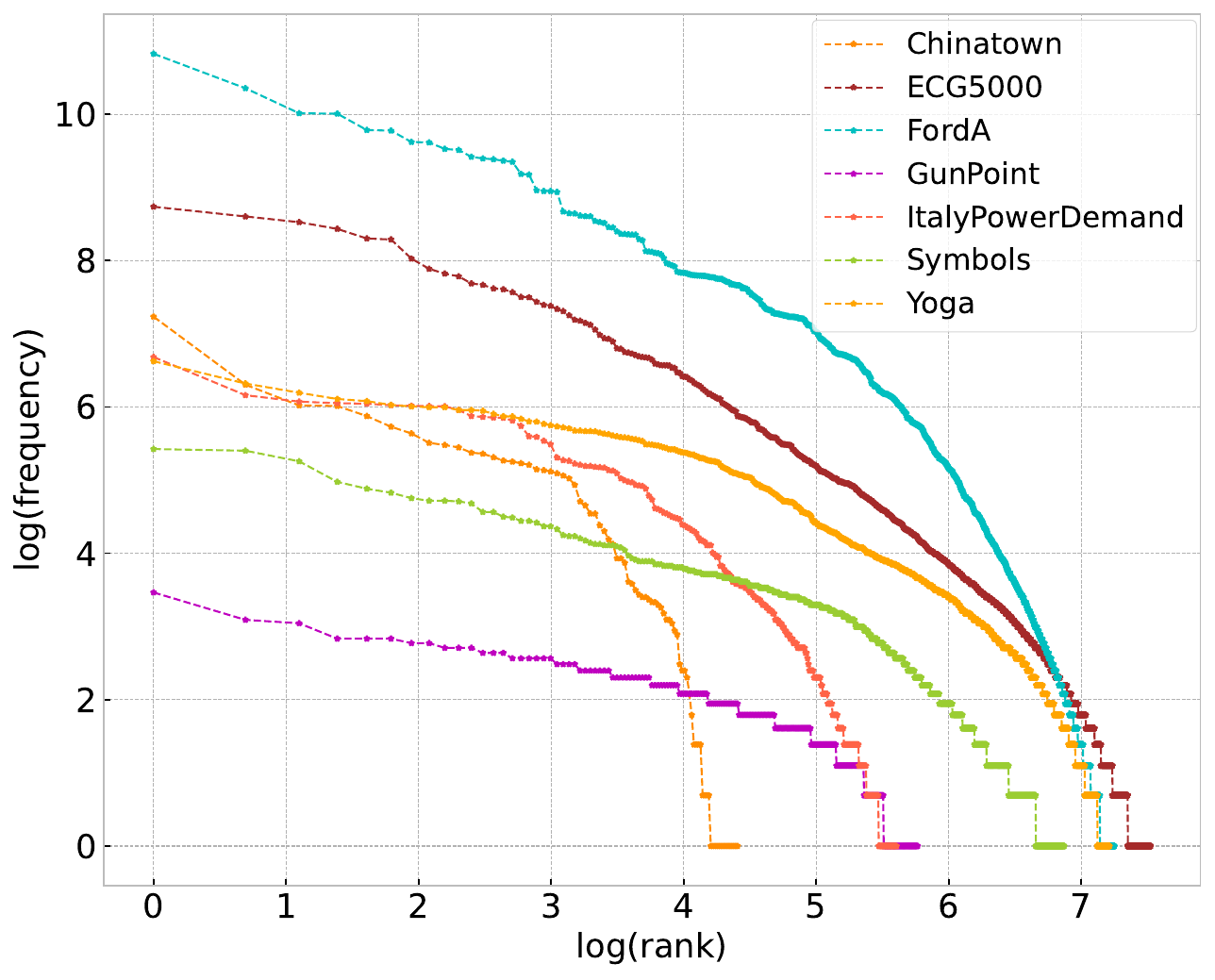}
\caption{Frequency and rank of symbols in various UCR datasets.}
\label{fig:zipf}
\end{figure}

In nearly all corpora, the most common word appears approximately twice as frequently as the next common word; this phenomenon is explained by Zipf's law \cite{powers-1998-applications}. Zipf's law asserts that the frequencies of certain events are inversely proportional to their rank, and further, the rank-frequency distribution is an inverse power law relation. In \figurename~\ref{fig:zipf}, we can see unigrams generated by ABBA symbolization from 7 different time series datasets from the UCR Archive coarsely meet Zipf's law. This showcases an appealing alignment between ABBA symbols and the native language words. 

\section{Experiments}\label{sec:exp}
In this section, we study three time series tasks to validate the efficiency of ABBA in LLM. We also fine-tune three language models on the training data using QLoRA \cite{dettmers2024qlora} with inhibition \cite{kang2024ina}. All experiments are simulated in PyTorch with a single NVIDIA A100 40GB GPU. The benefits of LLM-ABBA include (1) avoiding the need for LLMs to learn time series from scratch, and (2) only utilizing compression and decompression without the need for the training of extra embedding layers \cite{jin2023time}. All LLMs are quantized by 4-bits, and the used optimization method is adamw\_8bit. Each data set in the corresponding tasks is fine-tuned using fewer than 10 epoches. For a fair comparison, we evaluate our models on the same settings for each task. In the following, unless otherwise stated, we assume that the sorting-based aggregation is used for the ABBA digitization.

A larger dataset needs more symbols or LLM tokens, as a larger time series dataset contains more information and symbolic semantics. RoBERTa$_{\texttt{Large}}$ considers two directions of the input language sentence, and Llama-2-7B and Mistral-7B originate from the GPT architecture \cite{radford2019language} that only takes one direction (from left to right) into account. Causality analysis which is frequently used to compute the contextual of each signal has been widely used to analyze multichannel EEG signals. However, ECG signals mostly rely on sequential features. Thus, we infer that when using LLM-ABBA to analyze medical time series, the properties and characteristics should be analyzed first. For some datasets, we could not find or reproduce SOTA performance numbers. For a comprehensive analysis, we test ABBA with LLMs on three main time series analysis tasks. In this section, three LLMs are used to process the COPs in symbolic series, including RoBERTa$_{\texttt{Large}}$~\cite{liu2019roberta}, Llama-2-7B~\cite{touvron2023llama}, and Mistral-7B~\cite{jiang2023mistral}.

\subsection{Hyperparameters}

\subsubsection{Hyperparameters of ABBA}

There are four interactive parameters that establish the transition of time series when integrating ABBA into LLMs. The tolerance $\tol$ is chosen from $ \{1 \times 10^{-2}, 1 \times 10^{-4}, 1 \times 10^{-6}\}$ to control the degree of the compression and dimension reduction, and the digitization parameter $\alpha$ is chosen from $\{1 \times 10^{-2}, 1 \times 10^{-4}, 1 \times 10^{-6}\}$ to determine the number of distinct symbols.  $\mathcal{L}$ is a finite letter set that can be specified as the LLMs' tokens, and $\texttt{scl} \in \{1, 2, 3\}$ is used as the normalized scaling for the length of each piece.

\subsubsection{Hyperparameters of LLMs}

\begin{table}[hbt!]
\centering
\caption{Hyperparameters of classification tasks. Quant. is the model quantization process. Inhib. is the inhibition threshold in LoRA. Embed. means to save tuned embeddings. Optims. is the optimization method. LR is the learning rate.}
\label{table:Hyperclassification}
\setlength\tabcolsep{1pt}
\begin{tabular}{lcccccccccccc}
\hline
\multicolumn{9}{c}{\textbf{LLM-ABBA on Classification Tasks}}                                                                                                                       \\ \hline
\multirow{2}{*}{Models}    & Tokens & \multicolumn{5}{c}{LoRA}                        & \multirow{2}{*}{LR} & \multirow{2}{*}{\begin{tabular}[c]{@{}c@{}}Batch\\ Size\end{tabular}} \\ \cline{2-7}
                           & Length & alpha & $r$         & dropout & inhib. & Embed. &                     &                                                                       \\ \hline
RoBERTa$_{\texttt{Large}}$ & 512    & 16    & 16, 64, 256 & 0.05    & 0.3    & Save   & 5e-7                & 4                                                                     \\
Llama-2-7B                  & 4,096  & 16    & 16, 64, 256 & 0.05    & 0.3    & Save   & 5e-7                & 4                                                                     \\
Mistral-7B                 & 4,096  & 16    & 16, 64, 256 & 0.05    & 0.3    & Save   & 5e-7                & 4                                                                     \\ \hline
\end{tabular}
\end{table}

\begin{table}[hbt!]
\centering
\caption{Hyperparameters of regression and forecasting tasks. Quant. is the model quantization process. Inhib. is the inhibition threshold in LoRA. Embed. means to save tuned embeddings. Optims. is the optimization method.}
\label{table:Hyperregression}
\setlength\tabcolsep{1pt}
\begin{tabular}{lcccccccccccc}
\hline
\multicolumn{9}{c}{\textbf{LLM-ABBA on Regression and Forecasting Tasks}}                                                                                                                       \\ \hline
\multirow{2}{*}{Models}    & Tokens & \multicolumn{5}{c}{LoRA}                        & \multirow{2}{*}{LR} & \multirow{2}{*}{\begin{tabular}[c]{@{}c@{}}Batch\\ Size\end{tabular}} \\ \cline{2-7}
                           & Length & alpha & $r$         & dropout & inhib. & Embed. &                     &                                                                       \\ \hline
RoBERTa$_{\texttt{Large}}$ & 512    & 16    & 16, 64, 256 & 0.05    & 0.3    & Save   & 2e-6                & 4                                                                     \\
Llama-2-7B                  & 4,096  & 16    & 16, 64, 256 & 0.05    & 0.3    & Save   & 2e-6                & 4                                                                     \\
Mistral-7B                 & 4,096  & 16    & 16, 64, 256 & 0.05    & 0.3    & Save   & 2e-6                & 4                                                                     \\ \hline
\end{tabular}
\end{table}

There are three time series analysis tasks: classification, regression, and forecasting. We quantize LLMs by 4-bits using the bitsandbytes package\footnote{\url{https://github.com/bitsandbytes-foundation/bitsandbytes}}. In order to fine-tune LLMs, the shunting inhibition mechanism \cite{kang2024ina} is utilized during the QLoRA adapter fine-tuning progress. The modified embedding layer is also saved after fine-tuning on the corresponding task. For the classification task, the metric is accuracy rate (\%). Root-mean-square-error is used as the metric for regression tasks. Mean-square-error and mean-absolute-error are used as the metrics for forecasting tasks, and we also visualize the correlation coefficient of forecasting tasks on ETTh1 data in terms of their seven features. We control the fine-tuning epoch and apply a small batch size on every task. The alpha of QLoRA is set to 16. 

\subsection{Compression and recovery}

To transform the numerical time series to symbolic time series, we use tokens of LLMs as the initial dictionary of ABBA for the symbolic representation, and there are no extra tokens that will be used to represent the numerical input. ABBA shows a strong symbolic transition on time series signals (See \tablename~\ref{table:ABBA_performance}).

To visualize the performance of ABBA on time series transition processes, we employ ETTh1 time series data to compute the correlation coefficient and reconstruction error of ABBA. This multivariate data has seven features, and in terms of these seven features, the average of mean-absolute-error (MSE), mean-square-error (MAE), and correlation coefficient between original time series input and reconstructed outputs is computed.

\begin{table}[htp!]
\centering
\caption{Symbolic approximation performance on ETTh1 data using ABBA. ABBA describes a time series sample by using symbolic approximation, and the number of used symbols depdnds on the complexity of the data. If the time series sample is a regular wave (for example, a sine wave), the number of used symbols is small; otherwise, ABBA needs more symbols.}
\label{table:ABBA_performance}
\setlength\tabcolsep{0.5pt}
\begin{tabular}{cccccc}
\hline
\multicolumn{2}{c}{\textbf{Parameter Settings}}   & \textbf{\begin{tabular}[c]{@{}c@{}}Number of\\ Symbols\end{tabular}} & \multicolumn{3}{c}{\textbf{Reconstructed Time Series}}    \\ \hline
\begin{tabular}[c]{@{}c@{}}$\tol$ and 
 $\alpha$\end{tabular}     & \texttt{scl} & \begin{tabular}[c]{@{}c@{}}Used LLM's\\ tokens\end{tabular}& MSE & MAE & \begin{tabular}[c]{@{}c@{}}Correlation\\ Coefficient\end{tabular} \\ \hline
$1\times 10^{-2}$, $1\times 10^{-2}$      & 3   & 846    & $2.5 \times 10^{-7}$    & $1 \times 10^{-2}$    & $1.0$      \\
$1\times 10^{-4}$, $1\times 10^{-4}$      & 3   & 2,713   & $4.2\times 10^{-8}$    & $1.4 \times 10^{-4}$     & $1.0$      \\
$1\times 10^{-6}$, $1 \times 10^{-6}$      & 3   & 2,789   & $3.2 \times 10^{-8}$    & $1.3 \times 10^{-4}$    & 1.0      \\ \hline
\end{tabular}
\end{table}

In this section, we observe which ABBA settings better suit time series characteristics. The default $\texttt{scl}$ is set to $3$, which is used in other LLM tasks. $\tol$ and $\alpha$ are set to be the same. \tablename~\ref{table:ABBA_performance} reports the input-168-predict-96 results when using ABBA to reconstruct ETTh1 data in terms of seven features. Setting smaller $\tol$ and $\alpha$ in ABBA can reduce the MSE and MAE scores, but more symbols or LLM tokens will be used. Under all above conditions, the correlation coefficient is 1.0.

\subsection{Robustness of LLM-ABBA on MIT-BIH data}

\begin{figure}[htbp]
    \centering 
    \begin{subfigure}[b]{0.24\textwidth} 
        \centering 
        \includegraphics[scale=0.28]{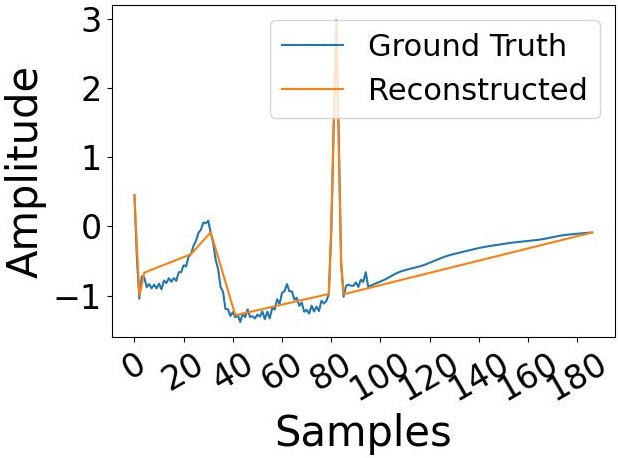} 
        \caption{\small $0 dB$ Gaussian noise.} 
    \end{subfigure} 
    \begin{subfigure}[b]{0.24\textwidth}
         \centering 
         \includegraphics[scale=0.28]{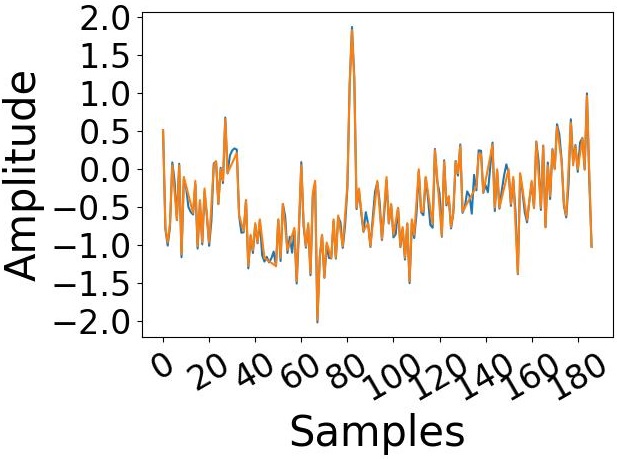} 
         \caption{\small $5 dB$ Gaussian noise.}  
    \end{subfigure}
        \begin{subfigure}[b]{0.24\textwidth}
         \centering 
         \includegraphics[scale=0.28]{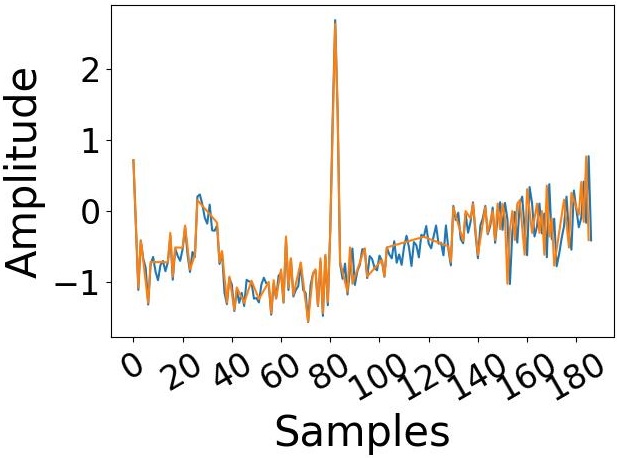} 
         \caption{\small $10 dB$ Gaussian noise.}  
    \end{subfigure}
        \begin{subfigure}[b]{0.24\textwidth}
         \centering 
         \includegraphics[scale=0.28]{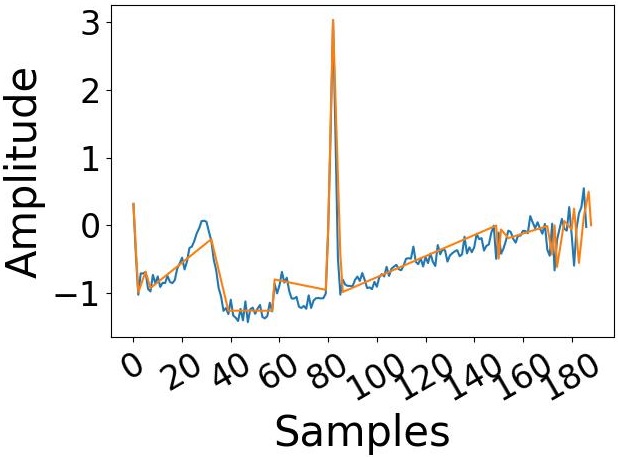} 
         \caption{\small $20 dB$ Gaussian noise.}  
    \end{subfigure}
    \caption{Visualization of MIT-BIH with Gaussian noise at SNR $\in$ $\{0 dB, 5 dB, 10 dB, 20 dB\}$ using ABBA.} 
    \label{figs:SNR} 
\end{figure}

\figurename~\ref{figs:SNR} presents the reconstruction of ABBA with FAPCA on MIT-BIH data that is added with Gaussian noise at signal-to-noise ratio (SNR) $\in$ $\{0 dB, 5 dB, 10 dB, 20 dB\}$. \tablename~\ref{table:SNR} shows the performance of Llama-2-7B-ABBA under different signal-to-noise ratio (SNR) cases. In the MIT-BIH data, the added Gaussian noise has a negligible impact on the performance of Llama-2-7B-ABBA.

\begin{table}[h]
\centering
\caption{The accuracy of Llama-2-ABBA on MIT-BIH data with additive Gaussian noise at SNR $\in$ $\{0 dB, 5 dB, 10 dB, 20 dB\}$ using Llama-2-7B.}
\label{table:SNR}
\setlength\tabcolsep{2.5pt}
\begin{tabular}{lccccccc}
\hline
\multirow{2}{*}{\textbf{SNR}} & \multicolumn{2}{c}{\textbf{Parameter Settings}} & \textbf{Classes} & \textbf{Symbols} & \multicolumn{3}{c}{\textbf{QLoRA}} \\ \cline{2-3} \cline{6-8} 
                              & $\tol$             & $\alpha$              & Number           & Number           & r = 16      & r = 64      & r = 256     \\ \hline
0 dB                          & 0.01               & 0.01                  & 5                & 4,578             & 89.2            & 89.0            & 88.6            \\
5 dB                          & 0.01               & 0.01                  & 5                & 4,610             & 87.2            & 87.1            & 86.9            \\
10 dB                         & 0.01               & 0.01                  & 5                & 5,391             & 90.0            & 89.7            & 88.7            \\
20 dB                         & 0.01               & 0.01                  & 5                & 10,642             & 89.3            & 88.9            & 88.7            \\ \hline
\end{tabular}
\end{table}

\subsection{Time series classification tasks}

For the classification task, we evaluate these three pretrained LLMs on UCR Time Series Archive datasets \cite{UCRArchive2018}, EEG eye state \cite{seyfi2022generating}, and MIT-BIH \cite{liu2021ecg} which have been extensively adopted for benchmarking time series classification models. We utilize cross-entropy loss for the classification training. Details of the implementation and datasets can be found in \tablename~\ref{table:Hyperclassification}. The evaluation metric is accuracy rate (\%). 

The UCR Archive contains 128 datasets already partitioned into train and test sets, although the ratio of the train set and test set is not always consistent\footnote{The UCR Archive 2018 is available at \url{https://www.cs.ucr.edu/~eamonn/time_series_data_2018/}.}. These datasets have varying numbers of labels and feature dimensions. Also, there can be uneven numbers of labels, which often results in overfitting. Therefore, classifying time series in the UCR Archive is a challenging task. \tablename~\ref{table:classificationAll} reports the full time series classification results on UCR2018. J1 refers to results of using k-means in the digitization process, and J2 refers to the results of using sorting-based aggregation in the digitization process. In practice, it is observed that sorting-based aggregation outperforms k-means for time series transition progress in most cases. 

\begin{table*}[ht]
\centering
\caption{Full comparison of results for time series classification tasks(\%) on UCR datasets.}
\label{table:classificationAll}
\setlength\tabcolsep{2.5pt}
\resizebox{0.90\linewidth}{!}{
\begin{tabular}{lcccccccccccccc}
\hline\scriptsize
\multirow{2}{*}{\textbf{Data}} & \textbf{Classes} & \textbf{Symbols} & \multicolumn{3}{c}{\textbf{RoBERTa$_{\texttt{Large}}$-ABBA}} & \multicolumn{3}{c}{\textbf{Llama-2-7B-ABBA}} & \multicolumn{3}{c}{\textbf{Mistreal-7B-ABBA}} & \multicolumn{2}{c}{\textbf{V2Sa \cite{yang2021voice2series}}} \\ \cline{4-14} 
 & Number  &  Number& Para.  & J1  & J2 & Para. & J1  & J2  & Para. & J1   & J2   & Para.   & SOTA    \\ \hline
Coffee     & 2 & 701 & 2.65M& 50.0 & 89.3  & 12.7M & 60.7& \textcolor{DarkGreen}{96.5}& 9.56M & 78.6 & 89.3  & 0.3M & \textbf{100}    \\
ECG200     & 2 & 1,781 & 2.65M& 70.0 & 68.0  & 12.7M & 63.0 & 64.0& 9.56M & 66.8  & 68.0  & 0.3M & \textbf{87.4}   \\
ECG5000    & 5 & 10,334 & 2.65M& \textcolor{DarkGreen}{81.2} & 76.0  & 12.7M & 75.7 & 74.7& 9.56M & 75.4  & 73.4  & 0.3M   & \textbf{94.0}   \\
Earthquakes& 2 & 940 & 2.65M& 52.7 & 74.8  & 12.7M & 77.7 & 76.3& 9.56M & \textbf{79.1}  & 76.3  & 0.3M   & \textcolor{DarkGreen}{78.4}   \\
FordA      & 2 & 9,759 & 2.65M& \textcolor{DarkGreen}{68.9} & 68.9  & 12.7M & 58.7 & 61.1& 9.56M & 62.7  & 60.9  & 0.3M   & \textbf{100}    \\
FordB      & 2 & 9,352 & 2.65M& \textcolor{DarkGreen}{68.9} & 58.1  & 12.7M & 56.1 & 58.9& 9.56M & 55.1  & 57.0  & 0.3M   & \textbf{100}    \\
GunPoint   & 2 & 791 & 2.65M& 51.4 & 73.3  & 12.7M & 54.0 & \textcolor{DarkGreen}{82.7}& 9.56M & 48.0  & 80.0  & 0.3M    & \textbf{96.7}     \\
HandOutlines  & 2 & 7,572 & 2.65M& 66.5 & \textcolor{DarkGreen}{77.0}  & 12.7M & 63.5 & 68.6& 9.56M & 65.1  & 71.6  & 0.3M   & \textbf{93.2}   \\
Herring    & 2 & 982 & 2.65M& 59.4 & \textcolor{DarkGreen}{65.6}  & 12.7M & 62.5 & 62.5& 9.56M & 54.7  & 60.9& 0.3M    & \textbf{68.8}   \\
ItalyPowerDemand     & 2 & 1,759 & 2.65M& 59.7 & 70.4  & 12.7M & 55.7 & \textcolor{DarkGreen}{73.4} & 9.56M & 53.4  & 73.2& 0.3M    & \textbf{97.1}   \\
Lightning2 & 2 & 2,175 & 2.65M& 67.2 & 65.6  & 12.7M & \textcolor{DarkGreen}{68.9} & 65.6& 9.56M & 67.2  & 62.3  & 0.3M   & \textbf{100}    \\
MiddlePhalanxOutlineCorrect    & 2 & 1,700 & 2.65M& 59.8 & 67.4  & 12.7M & 58.1 & 69.8& 9.56M & 61.2  & \textcolor{DarkGreen}{67.7} & 0.3M    & \textbf{91.1}   \\
MiddlePhalanxTW      & 6 & 1345 & 2.65M& 53.9 & \textcolor{DarkGreen}{54.5}  & 12.7M & 53.9 & 48.7& 9.56M & 51.9  & 46.8& 0.3M    & \textbf{84.9}   \\
SmallKitchenAppliances  & 2 & 2,207 & 2.65M& 66.2 & \textcolor{DarkGreen}{69.3}  & 12.7M & 60.8 & 63.2& 9.56M & 57.6  & 61.6& 0.3M    & \textbf{83.5}   \\
Strawberry & 2 & 3,593 & 2.65M& 71.2 & 85.1  & 12.7M & 69.5 & 84.9& 9.56M & 69.5  & \textcolor{DarkGreen}{88.4}& 0.3M    & \textbf{97.6}   \\
Trace      & 4 & 870 & 2.65M& 49.5 & 88.0  & 12.7M & 54.0 & \textcolor{DarkGreen}{90.0} & 9.56M & 47.0  & 77.0  & 0.3M   & \textbf{100}    \\
TwoLeadECG & 2 & 2,487 & 2.65M& 59.6 & 69.1  & 12.7M & 53.2 & 64.6& 9.56M & 53.2  & 63.9  & 0.3M   & \textbf{97.8}   \\
Wafer      & 2 & 4,805 & 2.65M& 94.6 & \textcolor{DarkGreen}{96.8}  & 12.7M & 91.3 & 93.5& 9.56M & 90.9  & 95.2  & 0.3M   & \textbf{100}    \\
Wine& 2 & 171 & 2.65M& 53.6 & 57.4  & 12.7M & 59.3 & 63.0& 9.56M & 63.0  & 55.6  & 0.3M   & \textbf{90.7}   \\
Worms      & 5 & 5,377 & 2.65M& 62.6 & \textcolor{DarkGreen}{67.5}  & 12.7M & 57.1 & 64.9& 9.56M & 54.5  & 63.6& 0.3M    & \textbf{83.1}   \\
WormsTwoClass & 2 & 5377 & 2.65M& 74.3 & \textcolor{DarkGreen}{81.8}  & 12.7M & 62.3 & 70.1& 9.56M & 61.0  & 79.2& 0.3M    & \textbf{98.7}   \\ \hline
\end{tabular}}
\end{table*}

In \tablename~\ref{table:classificationAll}, we report the classification performance on a partial dataset of UCR2018. In most cases, although LLM-ABBA cannot outperform the SOTA in terms of time series classification tasks, ABBA with LLMs can reach an acceptable application requirement in some practical cases (such as ``Coffee'', ``Earthquakes'', ``Herring'', ``Strawberry'', ``Trace'', ``Wafer'', ``WormsTwoClass''). Compared to V2S \cite{yang2021voice2series} which is the SOTA, although these three LLMs with the use of QLoRA occupies more memory, the multi-modality of LLMs, especially on time series analysis tasks, achieves a noticeable improvement.

In the medical domain (for example, identifying the eye state using EEG signals, distinguishing abnormal ECG signals, and classifying the ``normal beats", ``supraventricular ectopy beats", ``ventricular ectopy beats", ``fusion beats", and ``unclassifiable beats" of ECG signals), we report the performance of LLM-ABBA on three medical time series datasets. We set $\tol=\alpha=0.01$. In \tablename~\ref{table:medicalAll}, compared to CNN \cite{Kachuee2018ECGHC} on the PTB-DB data set, LLM-ABBA achieves performance almost equivalent to the SOTA. In the aspect of distinguishing MIT-BIH, CNN \cite{Kachuee2018ECGHC} and BiRNN \cite{
shashikumar2018detection} performs the best, but LLM-ABBA slightly outperforms LSTM \cite{saadatnejad2019lstm}.

In this section, our primary contribution is not to achieve SOTA accuracy but to pioneer a new paradigm: enabling LLMs to natively understand time series via ABBA, using only a single consumer-grade GPU for fine-tuning. We realize that our proposed method shows only a small performance gap compared to the latest state-of-the-art results reported in Middlehurst et al. \cite{Middlehurst2024}. Nevertheless, it delivers highly competitive results that are within a narrow margin of the current best, thereby bridging time series classification via symbolic approximation with LLMs’ capability to capture COPs.

\begin{table*}[h]
\centering
\caption{Full comparison of results on medical time series classification tasks(\%) on EEG eye states, ptb-db, and MIT-BIH.}
\label{table:medicalAll}
\setlength\tabcolsep{2.5pt}
\begin{tabular}{lcccccccccccccc}
\hline\scriptsize
\multirow{2}{*}{\textbf{Data}} & \multicolumn{1}{c}{\multirow{1}{*}{\textbf{Classes}}}  &\multicolumn{1}{c}{\multirow{1}{*}{\textbf{Symbols}}}    & \multicolumn{3}{c}{\textbf{RoBERTa$_{\texttt{Large}}$-ABBA}} & \multicolumn{3}{c}{\textbf{Llama-2-7B-ABBA}} & \multicolumn{3}{c}{\textbf{Mistreal-7B-ABBA}} & \multirow{2}{*}{\begin{tabular}[c]{@{}c@{}}\textbf{CNN} \\ \cite{Kachuee2018ECGHC} \end{tabular}}  & \multirow{2}{*}{\begin{tabular}[c]{@{}c@{}}\textbf{BiRNN} \\ \cite{shashikumar2018detection} \end{tabular}}  & \multirow{2}{*}{\begin{tabular}[c]{@{}c@{}}\textbf{LSTM}  \\ \cite{saadatnejad2019lstm} \end{tabular}} \\ \cline{4-12}
 & Number   & Number   & r=16  & r=64  & r=256 & r=16 & r=64& r=256& r=16 & r=64 & r=256 &  \\ \hline
EEG & 2  & 938  & 60.1      & \textbf{66.0}      & 64.4& 55.9     & 57.4     & 57.5      & 58.5      & 58.0     & 60.1     & 53.1   & 55.3      & 50.7 \\
ptb-db      & 2   & 2,179 & 89.5      & 90.6      & 89.3& \textcolor{DarkGreen}{99.0}     & 98.6     & 98.3      & 98.9      & 98.7     & 98.6  & \textbf{99.4}      & 97.0      & 90.7 \\
mit-bih     & 5    & 4,578      & 86.4      & 86.4      & 86.3& 89.6     & 89.4     & 89.1      & 89.3      & 89.7     & 89.3     & \textcolor{DarkGreen}{93.4}     & \textbf{96.5}     & 88.1  \\ \hline
\end{tabular}
\end{table*}

\subsection{Time series regression tasks}

\begin{table*}[h]
\centering
\caption{Full comparison of results on the regression task on 19 Monash Time Series Regression datasets.}
\label{table:regressionAll}
\setlength\tabcolsep{2.5pt}
\resizebox{0.98\linewidth}{!}{
\begin{tabular}{lcccccccccccc}
\hline\scriptsize
\multirow{2}{*}{\textbf{Data}} & \multirow{2}{*}{\textbf{Symbols}}  & \multicolumn{3}{c}{\textbf{RoBERTa$_{\texttt{Large}}$-ABBA}} & \multicolumn{3}{c}{\textbf{Llama-2-7B-ABBA}} & \multicolumn{3}{c}{\textbf{Mistreal-7B-ABBA}} & \multirow{2}{*}{\begin{tabular}[c]{@{}c@{}} \textbf{DrCIF}\\ \cite{GuijoRubio2024}\end{tabular}}  & \multirow{2}{*}{\begin{tabular}[c]{@{}c@{}}\textbf{FreshPRINCE} \\ \cite{middlehurst2022freshprince}\end{tabular}}\\ \cline{3-11}
&                                   & r=16                     & r=64                    & r=256                  & r=16                & r=64                & r=256               & r=16   & r=64           & r=256                                            & \multicolumn{1}{c}{}                                                                                                 & \multicolumn{1}{c}{}                                                                                                        \\ \cline{3-13} 
& Number                            & RMSE                     & RMSE                    & RMSE                   & RMSE                & RMSE                & RMSE                & RMSE   & RMSE           & RMSE                                             & \multicolumn{1}{c}{RMSE}                                                                                             & \multicolumn{1}{c}{RMSE}                                                                                                    \\ \hline
AppliancesEnergy                                & 778                               & \textbf{1.73}            & 2.09                    & 1.74                   & 2.43                & 2.43                & 2.43                & 2.34   & 2.02           & 2.11                                             & 2.81                                                                                                              & 2.27                                                                                                                 \\
HouseholdPowerConsumption1                      & 1717                              & 377.02                   & 377.20                  & 377.20                 & 398.01              & 398.05              & 398.05              & 228.83 & 228.78         & \textcolor{DarkGreen}{228.67} & 188.59                                                                                                   & \textbf{146.96}                                                                                                                  \\
HouseholdPowerConsumption2                      & 1717                              & 27.64                    & 27.71                   & 27.73                  & 36.63               & 36.71               & 36.69               & 24.54  & 24.56          & \textbf{24.51}                                   & 35.99                                                                                                              & 35.61                                                                                                                 \\
\hdashline 
BenzeneConcentration  & 3037                              & 4.01                     & \textbf{4.00}           & 4.00                   & 5.57                & 5.56                & 5.56                & 4.03   & 4.03           & 4.03                                             & 5.63                                                                                                             & 3.57                                                                                                                  \\
BeijingPM10Quality                              & 970                               & 66.16                    & 66.07                   & 66.07                  & 93.25               & 93.26               & 93.26               & 65.25  & 65.25          & \textbf{65.24}                                   & 96.49                                                                                                              & 93.05                                                                                                                \\
BeijingPM25Quality                              & 970                               & 54.16                    & 54.16                   & 54.16                  & 76.75               & 76.73               & 76.73               & 53.50  & \textbf{53.49} & 53.49                                            & 64.01                                                                                                                & 60.44                                                                                                                 \\
LiveFuelMoistureContent                         & 5689                              & \textbf{20.56}           & 20.56                   & 20.56                  & 29.32               & 29.33               & 29.32               & 20.94  & 20.88          & 20.85                                            & 34.92                                                                                                             & 33.69                                                                                                                \\
FloodModeling1                                  & 969                               & \textbf{0.002}            & 0.002                    & 0.002                   & 0.05                & 0.05                & 0.05                & 0.37   & 0.36           & 0.36                                             & 0.01                                                                                                              & 0.004                                                                                                                 \\
FloodModeling2                                  & 979                               & \textbf{0.002}            & 0.003                    & 0.002                   & 0.05                & 0.04                & 0.04                & 0.40   & 0.39           & 0.39                                             & 0.01                                                                                                             & 0.01                                                                                                                 \\
FloodModeling3                                  & 948                               & \textbf{0.002}            & 0.002                    & 0.002                   & 0.06                & 0.05                & 0.05                & 0.41   & 0.37           & 0.39                                             & 0.01                                                                                                             & 0.01                                                                                                                 \\
AustraliaRainfall                               & 4740                              & 4.36                     & 4.36                    & 4.36                   & 6.05                & 6.01                & 6.02                & 4.31   & \textbf{4.28}  & 4.30                                             & \multicolumn{1}{c}{}                                                                                                 & \multicolumn{1}{c}{}                                                                                                        \\ \hdashline 
PPGDalia              & 12298                             & 9.32                     & 9.32                    & 9.32                   & 12.54               & 12.50               & 12.52               & 9.04   & \textbf{9.02}  & 9.03                                             & 15.09                                                                                                             & 15.24                                                                                                                  \\
IEEEPPG                                         & 8971                              & 17.06                    & \textbf{17.00}          & 17.04                  & 22.59               & 22.53               & 22.55               & 17.15  & 17.12          & 17.16                                            & 34.92                                                                                                             & 33.69                                                                                                                \\
BIDMC32HR                                       & 9423                              & 6.73                     & 6.98                    & \textbf{6.71}          & 12.02               & 11.98               & 12.04               & 8.24   & 8.21           & 8.23                                             & 8.01                                                                                                             & 10.62                                                                                                               \\
BIDMC32RR                                       & 9412                              & 1.77                     & \textbf{1.74}           & 1.76                   & 2.64                & 2.61                & 2.62                & 2.09   & 2.06           & 2.08                                             & 4.30                                                                                                             & 4.04                                                                                                                \\
BIDMC32SpO2                                     & 5537                              & 2.90                     & \textbf{2.85}           & 2.89                   & 3.82                & 3.79                & 3.81                & 2.95   & 2.91           & 2.93                                             & 4.86                                                                                                             & 5.10                                                                                                               \\
\hdashline 
NewsHeadlineSentiment & 5537                              & \textbf{0.07}            & 0.07                    & 0.07                   & 0.13                & 0.13                & 0.13                & 0.11   & 0.11           & 0.11                                             & 0.15                                                                                                            & 0.15                                                                                                                \\
NewsTitleSentiment                              & 5537                              & \textbf{0.07}            & 0.07                    & 0.07                   & 0.13                & 0.13                & 0.13                & 0.11   & 0.11           & 0.11                                             & 0.14                                                                                                             & 0.14                                                                                                                 \\
\hdashline 
Covid3Month           & 227                               & \textbf{0.02}            & 0.02                    & 0.02                   & 0.11                & 0.11                & 0.11                & 0.45   & 0.44           & 0.44                                             & 0.04                                                                                                            & 0.04     \\ \hline
\end{tabular}
}
\end{table*}

For the regression task, we evaluate these three pretrained LLMs on the Time Series Extrinsic Regression (TSER) benchmarking archive  \cite{tan2021time}, which contains 19 time series datasets from 5 application domains, including Health Monitoring, Energy Monitoring, Environment Monitoring, Sentiment Analysis, and Forecasting\footnote{Monash regression data is available at \url{http://tseregression.org/}.}. To use as few symbols as possible, we initialize the setting of $\tol=0.01$ and $\alpha=0.01$. We also utilize the L2 loss for the regression training. Details of the implementation and datasets can be found in \tablename~\ref{table:Hyperregression}. The evaluation metric is root-mean-square-error (RMSE).

\begin{figure*}[htbp]
\centering
\includegraphics[width=1.0\textwidth]{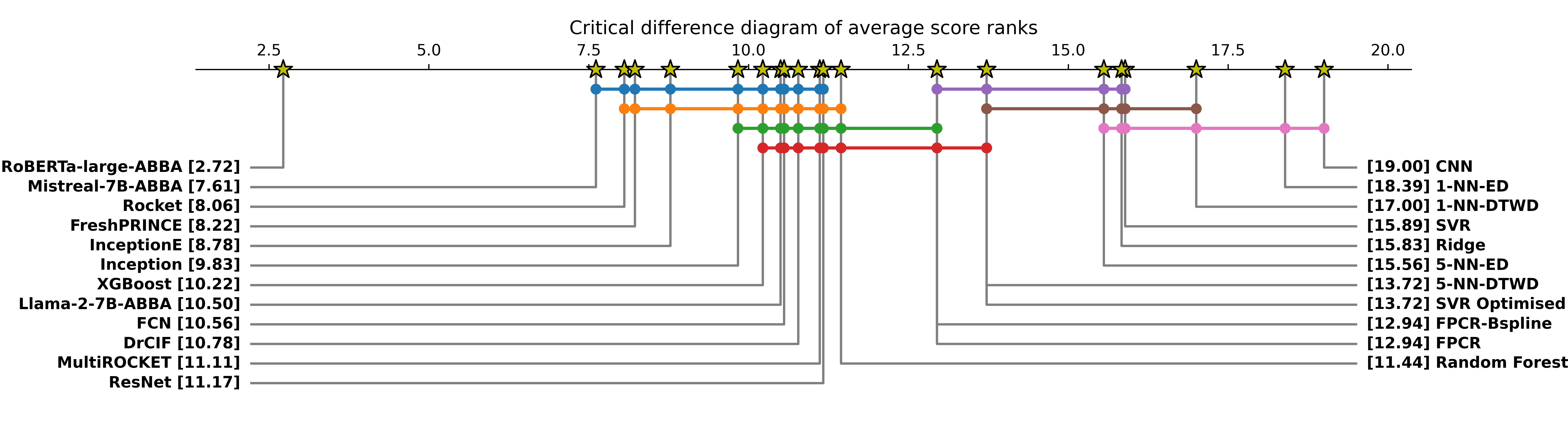}
\caption{The RMSE ranks for 23 Regressors on 19 TSER datasets. There are 20 of them used in \cite{GuijoRubio2024}.}
\label{fig:regression_CD}
\end{figure*}

Experimenting on the TSER benchmark archive \cite{tan2021time}, the empirical results are shown in \tablename~\ref{table:regressionAll}, in which for 12 out of 19 use-cases, RoBERTa$_{large}$-ABBA outperforms the machine learning SOTA results, such as DrCIF~\cite{GuijoRubio2024} and FreshPRINCE~\cite{middlehurst2022freshprince}. We believe that LLM-ABBA can exploit the semantic information hiding beneath the time series in the task of time series regression. Using the same post-hoc two-tailed Nemenyi test to \cite{tan2021time}, \figurename~\ref{fig:regression_CD} shows that ReBERTa$_{large}$-ABBA is the most accurate algorithm with an average rank of $2.72$. The critical difference (CD) is $7.356$. There are 23 regressors, and 20 of them come from \cite{GuijoRubio2024}.  ABBA is able to provide COPs to LLMs by compressing and digitizing time series to symbols, which finally results in the change of embedding space by using adaption fine-tuning methods.

\subsection{Time series forecasting tasks}

\begin{figure*}[htbp]
    \centering 
    \begin{subfigure}[b]{0.24\textwidth} 
 \centering 
 \includegraphics[scale=0.29]{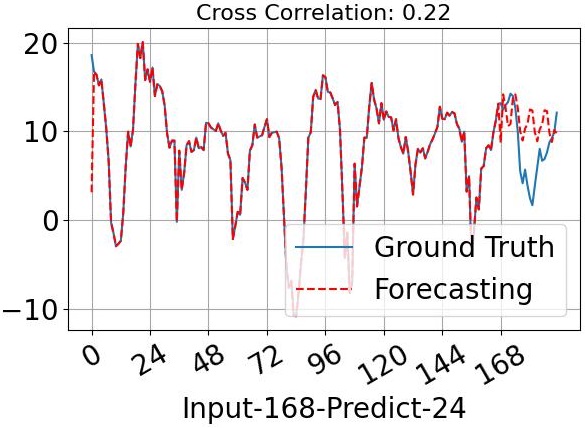} 
 \caption{\small Feature 1.} 
    \end{subfigure} 
    \begin{subfigure}[b]{0.24\textwidth}
 \centering 
 \includegraphics[scale=0.29]{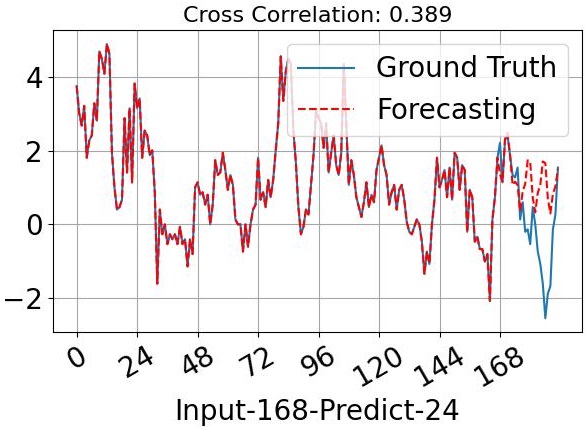} 
 \caption{\small Feature 2.} 
    \end{subfigure}
    \begin{subfigure}[b]{0.24\textwidth}
 \centering 
 \includegraphics[scale=0.29]{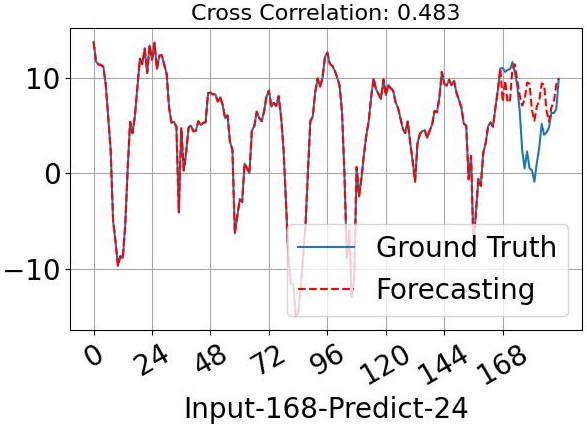} 
 \caption{\small  Feature 3.} 
    \end{subfigure}
    \begin{subfigure}[b]{0.24\textwidth}
 \centering 
 \includegraphics[scale=0.29]{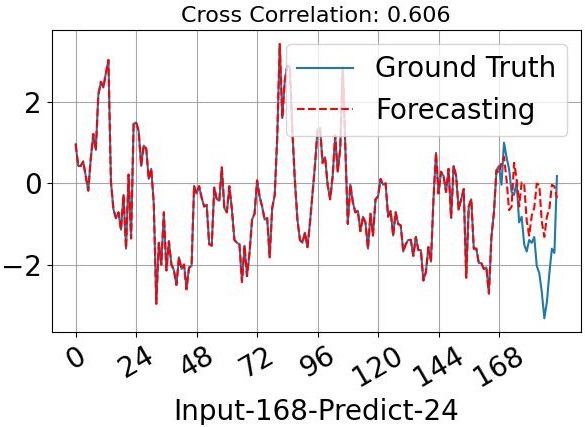} 
 \caption{\small  Feature 4.} 
    \end{subfigure}
    \begin{subfigure}[b]{0.32\textwidth}
 \centering 
 \includegraphics[scale=0.29]{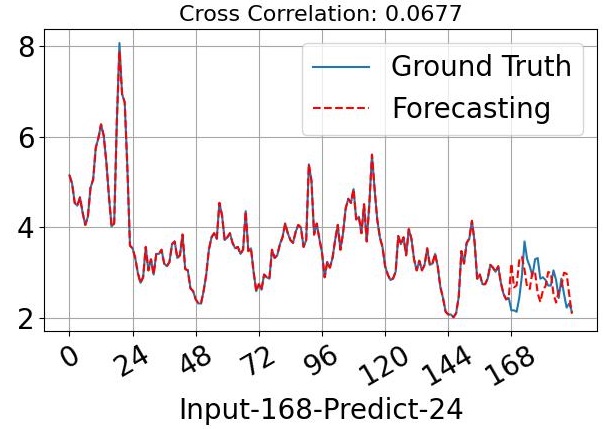} 
 \caption{\small  Feature 5.} 
    \end{subfigure}
    \begin{subfigure}[b]{0.32\textwidth}
 \centering 
 \includegraphics[scale=0.29]{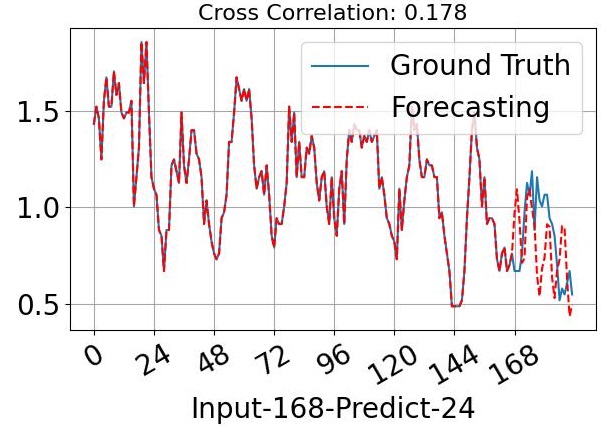} 
 \caption{\small  Feature 6.} 
    \end{subfigure}
 \begin{subfigure}[b]{0.32\textwidth}
 \centering 
 \includegraphics[scale=0.29]{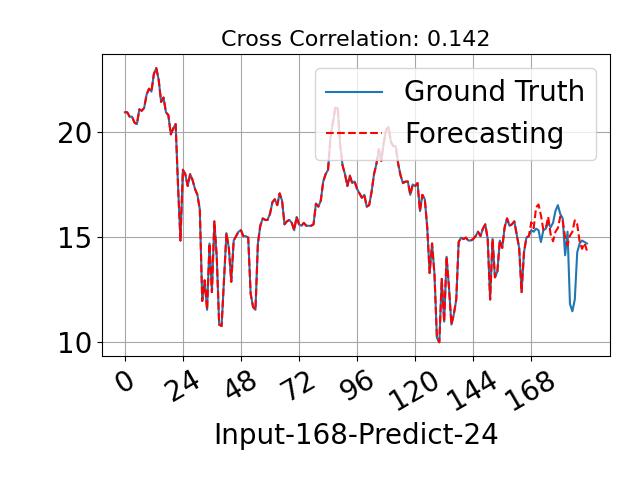} 
 \caption{\small Feature 7.} 
    \end{subfigure}
    \caption{Visualization of input-168-predict-24 results on ETTh1 using Mistreal-7B-ABBA.} 
    \label{figs: visualizeETTh1} 
\end{figure*}

\begin{table*}[h]
\centering
\caption{Full comparison of results for the forecasting task on 4 time series forecasting datasets.}
\label{table:forecastingAll}
\setlength\tabcolsep{2.2pt}
\begin{tabular}{lcccccccccccccccccccc}
\hline\scriptsize
\multirow{3}{*}{\textbf{Data}}   & \multirow{3}{*}{\textbf{Predictor}}    & \multirow{3}{*}{\textbf{Symbols}}    & \multicolumn{6}{c}{\textbf{Llama-2-7B-ABBA}}  & \multicolumn{6}{c}{\textbf{Mistreal-7B-ABBA}} & \multicolumn{2}{c}{\multirow{2}{*}{\begin{tabular}[c]{@{}c@{}} \textbf{Informer}\\ \cite{tan2021time}\end{tabular}}} & \multicolumn{2}{c}{\multirow{2}{*}{\begin{tabular}[c]{@{}c@{}} \textbf{Time-LLM}\\ \cite{jin2023time}\end{tabular}}} & \multicolumn{2}{c}{\multirow{2}{*}{\begin{tabular}[c]{@{}c@{}} \textbf{TimeMixer}\\ \cite{wang2024timemixer}\end{tabular}}} \\ \cline{4-15}
 &  &  & \multicolumn{2}{c}{r=16} & \multicolumn{2}{c}{r=64} & \multicolumn{2}{c}{r=256} & \multicolumn{2}{c}{r=16} & \multicolumn{2}{c}{r=64} & \multicolumn{2}{c}{r=256} \\ \cline{4-21} 
  & Length  & Number  & MSE  & MAE & MSE  & MAE & MSE& MAE & MSE  & MAE & MSE  & MAE & MSE  & MAE  & MSE  & MAE   & MSE  & MAE  & MSE  & MAE  \\ \hline
ETTh1 & 168/24 & 2,789 & 0.689 & 0.653  & 0.647 & 0.696  & 0.658  & 0.677 & 0.631 & 0.681  & 0.622 & 0.631  & 0.626  & 0.677  & 0.577 & 0.549  & -  & -  & -  & -\\
ETTh2 & 168/24   & 5,383   & 0.798  & 0.788 & 0.784  & 0.761 & 0.789   & 0.772 & 0.776  & 0.787 & 0.759  & 0.761 & 0.762  & 0.771  & 0.720 & 0.665   & -  & -  & -  & -\\
ETTm1 & 168/24   & 3,170   & 0.403  & 0.397 &  0.386 & 0.364 & 0.392   & 0.385 & 0.457  & 0.422 & 0.401  & 0.387 & 0.407  & 0.397   & 0.323  & 0.369   & -  & -  & -  & -\\
ETTm2 & 168/24    & 6,878 & 0.224  & 0.209 & 0.201 & 0.198   & 0.215 & 0.207  & 0.251 & 0.237  & 0.214 & 0.203  & 0.218  & 0.209 & -  & -   & -  & -  & -  & -\\  \hdashline
ETTh1 & 168/96    & 2,789  & 0.762 & 0.786  & 0.754 & 0.752   & 0.759 & 0. 60 & 0.792 & 0.804  & 0.773 & 0.782  & 0.7811  & 0.788  & -  & -      & 0.362  & 0.392    & 0.375  & 0.440 \\
ETTh2 & 168/96& 5,383  & 0.912 & 0.885  & 0.892 & 0.881   & 0.907 & 0.876  & 0.899 & 0.887  & 0.871 & 0.866  & 0.878  & 0.872  & -  & -   & 0.268  & 0.328  & 0.289  & 0.341\\
ETTm1 & 168/96& 3,170  & 0.542 & 0.537  & 0.531 & 0.528   & 0.538 & 0.520  & 0.541 & 0.533  & 0.524 & 0.517  & 0.529  & 0.520   & -  & -  & 0.272  & 0.233   & 0.320  & 0.357\\
ETTm2 & 168/96      & 6,878  & 0.302 & 0.286  & 0.288 & 0.267   & 0.293 & 0.278  & 0.289 & 0.302  & 0.276 & 0.281  & 0.280  & 0.285     & -  & -  & 0.161  & 0.253   & 0.175  & 0.258\\  \hdashline
ETTh1 & 168/168    & 2789  & 1.161 & 1.010  & 1.087 & 0.964   & 1.096 & 0.989  & 1.182 & 1.217  & 1.174 & 1.968  & 1.179  & 1.992    & 0.931  & 0.752   & 0.398  & 0.418  & 0.429  & 0.421\\
ETTh2 & 168/168& 5,383  & 4.103 & 2.675  & 3.975 & 2.101   & 4.086 & 2.537  & 4.092 & 2.626  & 3.898 & 2.134  & 3.910  & 2.245     & 3.489  & 1.515   & 0.329  & 0.375  & 0.372  & 0.392\\
ETTm1 & 168/168& 3,170  & 0.989 & 0.962  & 0.974 & 0.952   & 0.979 & 0.959  & 1.001 & 0.986  & 0.966 & 0.958  & 0.972  & 0.966       & 0.678  & 0.614   & 0.310  & 0.358  & 0.361  & 0.381\\
ETTm2 & 168/168      & 6,878  & 0.616 &  0.583 & 0.576 & 0.544   & 0.580 & 0.561  & 0.592 & 0.541  & 0.521 & 0.503  & 0.532  & 0.509    & -  & -    & 0.219  & 0.293   & 0.237  & 0.299\\  \hline
\end{tabular}

\end{table*}

For time series forecasting, we experimented on 4 well-established benchmarks: ETT datasets (including 4 subsets: ETTh1, ETTh2, ETTm1, ETTm2) \cite{zhou2021informer}. Details of the implementation and datasets can be found in \tablename~\ref{table:Hyperregression}. The input length of the time series is $168$, and we use three different forecasting horizons $H \in \{24, 96, 168\}$. The evaluation metrics include MSE and MAE.

Although LLM-ABBA cannot obtain a new SOTA on time series forecasting tasks, it compares favorably to the Informer architecture which is trained from scratch. The congenital defect of ABBA is that the symbolization tends to be affected by the fluctuation and oscillation of time series signals, which eventually leads to higher MSE and MAE scores. Because LLM-ABBA utilizes a totally different technical roadmap to existing methods, it only remolds the construction of the LLM's tokens. However, remodeling pretrained tokens inevitably brings the previous pretrained semantics to the LLM-ABBA design. Thus, we discussed the semantic consistency of LLM-ABBA using extra symbols or tokens to overcome this problem.

\subsection{QLoRA fine-tuning}

Because the low rank of adapter fine-tuning will influence the efficiency of passing information \cite{dettmers2024qlora,kang2024ina} from the previous layer, we use different low rank settings of QLoRA on the corresponding tasks during the fine-tuning progress. But for time series regression and forecasting tasks, we select $r \in \{16, 64, 256 \}$ for the corresponding data input. We find that there is no obvious over-fitting problem, and more tunable parameters are not able to improve the performance of LLM-ABBA. In medical time series domains, ptb-db and MIT-BIH arrhythmia data sets are mostly used. EEG eye state data set has two categories, and because of its high complexity, the accuracy always stays at around $60\%$. EEG eye state data and MIT-BIH has more than one channel, which indicates that LLM-ABBA might have the ability to process complicate features across channels. \tablename~\ref{table:medicalAll} presents the full medical time series classification results using LLM-ABBA.

LLM-ABBA achieves comparable time series forecasting results to the SOTAs, and there is no over-fitting in these tasks when using different low rank $r$. Because  ABBA tends to symbolize trends and altitudes of the time series signals, LLM-ABBA always strengthens the vibration of predicted time series segments which can be seen in \figurename~\ref{figs: visualizeETTh1}.

\subsection{FAPCA benefits the time series forecasting}

\begin{figure}[htbp]
    \centering 
    \begin{subfigure}[b]{0.24\textwidth} 
        \centering 
        \includegraphics[scale=0.28]{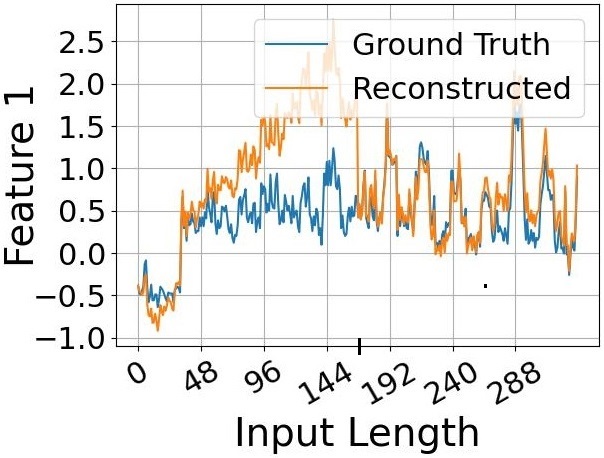} 
        \caption{\small Traditional ABBA.} 
    \end{subfigure} 
    \begin{subfigure}[b]{0.24\textwidth}
         \centering 
         \includegraphics[scale=0.28]{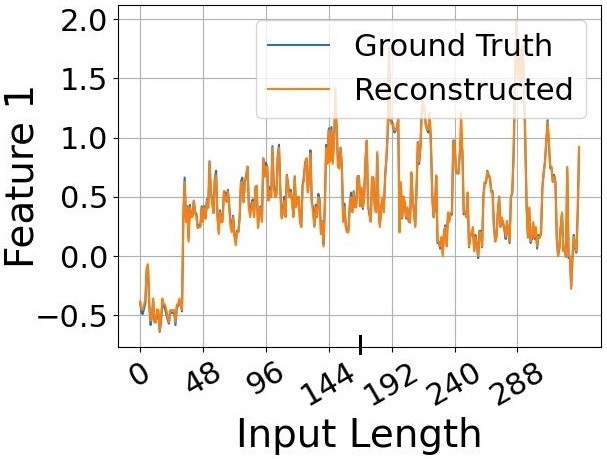} 
         \caption{\small ABBA with FAPCA.}  
    \end{subfigure}
    \caption{Visualization of Input-168-Predict-168 with the use of the traditional ABBA and the ABBA with FAPCA on ETTh1 data.} 
    \label{figs:JABBA_and_XABBA} 
\end{figure}

When focusing solely on the symbolic approximation in time series forecasting tasks, \figurename~\ref{figs:JABBA_and_XABBA} illustrates the difference between traditional ABBA and ABBA with FAPCA. 
If an incorrect replacement of a previous symbol occurs in the subsequent reconstruction, traditional ABBA is prone to the accumulation of errors. In contrast, FAPCA effectively mitigates this issue, particularly in long-term forecasting scenarios such as the Input-168-Predict-168 task.

\subsection{Semantic consistency}

When using pretrained tokens as the input symbols, fine-tuning on non-linguistic data (e.g., time series signals) may cause semantic drift in LLMs. To mitigate this issue, we construct a new set of symbols to extend ASCII codes by adding more digits and expanding the alphabet table to be used. Following the same fine-tuning configuration as in the previous experiments, we assess the forecasting performance by fine-tuning Mistral-7B. As shown in \tablename~\ref{table:SemanticLoss0}, the results are largely consistent with those in \tablename~\ref{table:forecastingAll}, indicating that the semantic loss is negligible.

\begin{table}[h]
\centering
\caption{The performance of LLM-ABBA with extra new tokens (symbolic ASCII codes) on ETTh1 data in terms of time series forecasting tasks.}
\label{table:SemanticLoss0}
\setlength\tabcolsep{1.7pt}
\begin{tabular}{lcccccccc}
\hline\scriptsize
\multirow{3}{*}{\textbf{Data}}   & \multirow{3}{*}{\textbf{Predictor}}    & \multirow{3}{*}{\textbf{Symbols}}   & \multicolumn{6}{c}{\textbf{Mistreal-7B-ABBA}} \\ \cline{4-9}
 &  &  & \multicolumn{2}{c}{r=16} & \multicolumn{2}{c}{r=64} & \multicolumn{2}{c}{r=256}  \\ \cline{4-9} 
  & Length  & Number  & MSE  & MAE & MSE  & MAE & MSE    & MAE  \\ \hline
ETTh1 & 168/24    & 2,789  & 0.636 & 0.692  & 0.626 & 0.632   & 0.629 & 0.681 \\
ETTh2 & 168/24& 5,383  & 0.779 & 0.788  & 0.761 & 0.763   & 0.763 & 0.777 \\
ETTm1 & 168/24& 3,170  & 0.457 & 0.402  & 0.402 & 0.387   & 0.407 & 0.399\\
ETTm2 & 168/24      & 6,878  & 0.253 & 0.238  & 0.215 & 0.203   & 0.219 & 0.209 \\ \hline
\end{tabular}

\end{table}

\subsection{Trade-off between runtime and performance}

\begin{figure}[htbp]
    \centering 
    \includegraphics[scale=0.3]{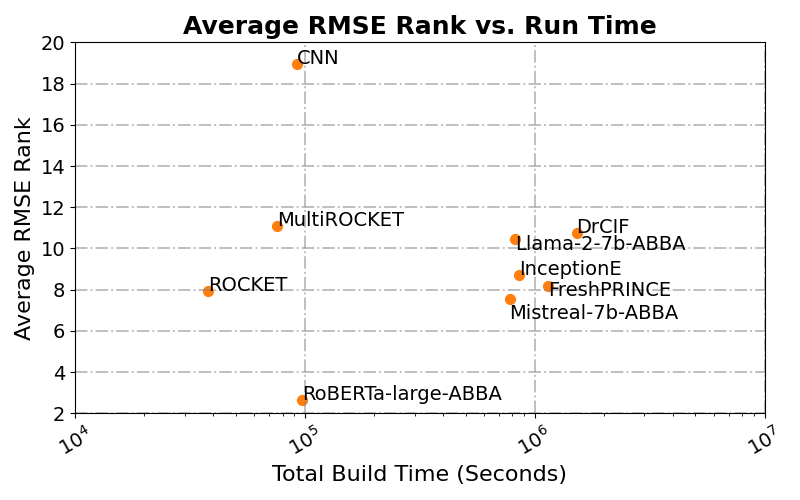} 
    \caption{Run time in seconds (log scale over 19 TSER datasets) plotted against average rank for RMSE.} 
    \label{figs:rmse_rank_run_time} 
\end{figure}

\figurename~\ref{figs:rmse_rank_run_time} shows the average rank RMSE against the run time (on a log scale) for the nine regressors: RoBERTa$_{large}$-ABBA, Mistreal-7B-ABBA, Llama-2-7b-ABBA, ROCKET \cite{dempster2020rocket}, MultiROCKET \cite{tan2022multirocket}, InceptionE \cite{ismail2020inceptiontime}, DrCIF \cite{GuijoRubio2024}, FreshPRINCE \cite{middlehurst2022freshprince}, and CNN \cite{ismail2019deep}. We
see a direct trade-off between runtime and performance. All algorithms run on a single thread CPU except for RoBERTa$_{large}$-ABBA, Mistreal-7B-ABBA, and Llama-2-7b-ABBA,
which ran on a GPU. The total run time of ReBERTa$_{large}$-ABBA on 19 TSER datasets is shorter than that of InceptionE \cite{ismail2020inceptiontime}, DrCIF \cite{GuijoRubio2024}, and FreshPRINCE \cite{middlehurst2022freshprince}. 
But because of the huge weight of LLMs, the learning time and the inference of LLM-ABBA are inevitably longer than that of other traditional machine learning methods in \cite{tan2021time,GuijoRubio2024}.

\section{Discussion}\label{sec:limit}

ABBA is assessed carefully via performance profiles with respect to its reconstruction evaluated via 2-norm, DTW, and their respective differenced measures, which show competitive performance against the SOTA STSA methods (e.g., SAX). Previous STSA methods have been applied in various data mining applications, e.g., EEG signal analysis \cite{Jain2024} and Internet of Things \cite{Jha2024}. ABBA also shows improved performance in the anomaly detection method TARZAN by simply replacing SAX methods \cite{EG19b, fABBA2022}.

LLMs can understand the generated symbols of ABBA. Each data sample can be approached by symbols, and each used symbol has a specific meaning that presents one node of the internal COPs of time series data. LLM-ABBA performs well not only on time series classification tasks, but also on time series regression tasks (as seen in \tablename~\ref{table:classificationAll} and \tablename~\ref{table:regressionAll}). Because these symbolic series have a logic chain that can represent the trend of time series samples, LLMs are able to learn the trend of time series via adapter fine-tuning methods. As shown in \figurename~\ref{figs: visualizeETTh1}, by using  the inverse symbolization process of ABBA, LLMs can predict the trend of time series signals, and these predicted parts have a smaller drift. Therefore, time series forecasting tasks also can demonstrate these findings. ABBA perfectly approximates time series via symbolic series, but because LLMs are born with hallucination, more generated contents would contain more ``hallucination knowledge''. Therefore, LLM-ABBA performs better on short-term time series forecasting tasks and time series regression tasks.

\section{Conclusion and Limitations}\label{sec:conclude}
In this paper, we propose LLM-ABBA for time series classification, regression, and forecasting tasks. We discuss how to seamlessly integrate time series symbolization with LLMs and enhance its performance. In theory, we analyze the reconstruction error of ABBA symbolization, how it relates to the dominant parameters, and the congenital defect of LLM-ABBA. To mitigate the drift phenomenon of time series, we introduce the FAPCA method to improve ABBA symbolization. The empirical results demonstrate our method achieves a performance comparable to the SOTA on classification and regression tasks. In terms of convenience and universality, LLM-ABBA improves the multi-modality of LLMs on time series analysis.

The proposed FAPCA strategy for ABBA cannot fully guarantee a complete removal of cumulative error arising from the previous mistaken symbols from the recovery. Additionally, the hallucination of LLMs cannot be addressed in this work, and the vibration or adverse response of predicted sequences can still have negative effects on final performance. Moreover, most LLMs only can support up to 4,096 tokens, which fundamentally prohibits long-term time series analysis tasks. Lastly, the learning time and the inference speed of LLM-ABBA are inevitably slower than most traditional machine learning methods.


\bibliographystyle{IEEEtran}
\bibliography{refs}

@inproceedings{middlehurst2022freshprince,
  title={The freshprince: A simple transformation based pipeline time series classifier},
  author={Middlehurst, Matthew and Bagnall, Anthony},
  booktitle={International Conference on Pattern Recognition and Artificial Intelligence},
  pages={150--161},
  year={2022},
  organization={Springer}
}

@article{tan2022multirocket,
  title={MultiRocket: multiple pooling operators and transformations for fast and effective time series classification},
  author={Tan, Chang Wei and Dempster, Angus and Bergmeir, Christoph and Webb, Geoffrey I},
  journal={Data Mining and Knowledge Discovery},
  volume={36},
  number={5},
  pages={1623--1646},
  year={2022},
  publisher={Springer}
}

@article{GuijoRubio2024,
  author    = {Guijo-Rubio, David and Middlehurst, Matthew and Arcencio, Gisele and others},
  title     = {Unsupervised feature based algorithms for time series extrinsic regression},
  journal   = {Data Mining and Knowledge Discovery},
  year      = {2024},
  volume    = {38},
  pages     = {2141--2185},
  doi       = {10.1007/s10618-024-01027-w},
  url       = {https://doi.org/10.1007/s10618-024-01027-w}
}

@article{ismail2020inceptiontime,
  title={Inceptiontime: Finding alexnet for time series classification},
  author={Ismail Fawaz, Hassan and Lucas, Benjamin and Forestier, Germain and Pelletier, Charlotte and Schmidt, Daniel F and Weber, Jonathan and Webb, Geoffrey I and Idoumghar, Lhassane and Muller, Pierre-Alain and Petitjean, Fran{\c{c}}ois},
  journal={Data Mining and Knowledge Discovery},
  volume={34},
  number={6},
  pages={1936--1962},
  year={2020},
  publisher={Springer}
}

@article{dempster2020rocket,
  title={ROCKET: exceptionally fast and accurate time series classification using random convolutional kernels},
  author={Dempster, Angus and Petitjean, Fran{\c{c}}ois and Webb, Geoffrey I},
  journal={Data Mining and Knowledge Discovery},
  volume={34},
  number={5},
  pages={1454--1495},
  year={2020},
  publisher={Springer}
}

@phdthesis{Chen2024thesis,
  author       = {Xinye Chen},
  title        = {Fast aggregation‐based algorithms for knowledge discovery},
  school       = {The University of Manchester},
  year         = {2024},
  month        = {August},
  type         = {PhD thesis},
  note         = {Department of Mathematics, supervised by Nicholas Higham \& Stefan G\"uttel},
  url          = {https://research.manchester.ac.uk/en/studentTheses/fast-aggregation-based-algorithms-for-knowledge-discovery}
}

@article{hu2022lora,
  title={{LoRA}: Low-rank adaptation of large language models.},
  author={Hu, Edward J and Shen, Yelong and Wallis, Phillip and Allen-Zhu, Zeyuan and Li, Yuanzhi and Wang, Shean and Wang, Lu and Chen, Weizhu and others},
  journal={ICLR},
  volume={1},
  number={2},
  pages={3},
  year={2022}
}

@article{baydogan2015learning,
  title={Learning a symbolic representation for multivariate time series classification},
  author={Baydogan, Mustafa Gokce and Runger, George},
  journal={Data Mining and Knowledge Discovery},
  volume={29},
  number={2},
  pages={400--422},
  year={2015},
  publisher={Springer}
}

@article{saadatnejad2019lstm,
  title={{LSTM}-based {ECG} classification for continuous monitoring on personal wearable devices},
  author={Saadatnejad, Saeed and Oveisi, Mohammadhosein and Hashemi, Matin},
  journal={IEEE journal of biomedical and health informatics},
  volume={24},
  number={2},
  pages={515--523},
  year={2019},
  publisher={IEEE}
}

@inproceedings{shashikumar2018detection,
  title={Detection of paroxysmal atrial fibrillation using attention-based bidirectional recurrent neural networks},
  author={Shashikumar, Supreeth P and Shah, Amit J and Clifford, Gari D and Nemati, Shamim},
  booktitle={Proceedings of the 24th ACM SIGKDD international conference on knowledge discovery \& data mining},
  pages={715--723},
  year={2018}
}

@article{Jha2024,
author={Jha, Vidyapati
and Tripathi, Priyanka},
title={Probabilistic {SAX}: A Cognitively-Inspired Method for Time Series Classification in Cognitive IoT Sensor Network},
journal={Mobile Networks and Applications},
year={2024}
}

@Article{Jain2024,
author={Jain, Divya
and Ranjan, Rakesh
and Sharma, Archana
and Sharma, Sanjaeev Narayan
and Jain, Alok},
title={Fast and accurate {ECG} signal peaks detection using symbolic aggregate approximation},
journal={Multimedia Tools and Applications},
year={2024},
month={Sep},
day={01},
volume={83},
number={30},
pages={75033-75059},
}

@article{radford2019language,
  title={Language models are unsupervised multitask learners},
  author={Radford, Alec and Wu, Jeffrey and Child, Rewon and Luan, David and Amodei, Dario and Sutskever, Ilya and others},
  journal={OpenAI blog},
  volume={1},
  number={8},
  pages={9},
  year={2019}
}

@article{Kachuee2018ECGHC,
  title={{ECG} Heartbeat Classification: A Deep Transferable Representation},
  author={Mohammad Kachuee and Shayan Fazeli and Majid Sarrafzadeh},
  journal={IEEE International Conference on Healthcare Informatics},
  year={2018},
  pages={443--444},
}

@article{cleveland1990stl,
  title={{STL}: A seasonal-trend decomposition},
  author={Cleveland, Robert B and Cleveland, William S and McRae, Jean E and Terpenning, Irma and others},
  journal={Journal of Official Statistics},
  volume={6},
  number={1},
  pages={3--73},
  year={1990}
}

@article{liu2024autotimes,
  title={{AutoTimes}: Autoregressive time series forecasters via large language models},
  author={Liu, Yong and Qin, Guo and Huang, Xiangdong and Wang, Jianmin and Long, Mingsheng},
  journal={arXiv preprint arXiv:2402.02370},
  year={2024}
}

@article{cao2023tempo,
  title={Tempo: Prompt-based generative pre-trained transformer for time series forecasting},
  author={Cao, Defu and Jia, Furong and Arik, Sercan O and Pfister, Tomas and Zheng, Yixiang and Ye, Wen and Liu, Yan},
  journal={arXiv preprint arXiv:2310.04948},
  year={2023}
}

@inproceedings{liu2024unitime,
  title={{UniTime}: A language-empowered unified model for cross-domain time series forecasting},
  author={Liu, Xu and Hu, Junfeng and Li, Yuan and Diao, Shizhe and Liang, Yuxuan and Hooi, Bryan and Zimmermann, Roger},
  booktitle={Proceedings of the ACM on Web Conference 2024},
  pages={4095--4106},
  year={2024}
}

@article{zhou2023one,
  title={One fits all: Power general time series analysis by pretrained lm},
  author={Zhou, Tian and Niu, Peisong and Sun, Liang and Jin, Rong and others},
  journal={Advances in neural information processing systems},
  volume={36},
  pages={43322--43355},
  year={2023}
}

@article{xue2023promptcast,
  title={{PromptCast}: A new prompt-based learning paradigm for time series forecasting},
  author={Xue, Hao and Salim, Flora D},
  journal={IEEE Transactions on Knowledge and Data Engineering},
  year={2023},
  publisher={IEEE}
}

@article{ismail2019deep,
  title={Deep learning for time series classification: a review},
  author={Ismail Fawaz, Hassan and Forestier, Germain and Weber, Jonathan and Idoumghar, Lhassane and Muller, Pierre-Alain},
  journal={Data Mining and Knowledge Discovery},
  volume={33},
  number={4},
  pages={917--963},
  year={2019},
  publisher={Springer}
}

@article{ismail2020benchmarking,
  title={Benchmarking deep learning interpretability in time series predictions},
  author={Ismail, Aya Abdelsalam and Gunady, Mohamed and Corrada Bravo, Hector and Feizi, Soheil},
  journal={Advances in neural information processing systems},
  volume={33},
  pages={6441--6452},
  year={2020}
}

@inproceedings{
jin2023time,
title={Time-{LLM}: Time Series Forecasting by Reprogramming Large Language Models},
author={Ming Jin and Shiyu Wang and Lintao Ma and Zhixuan Chu and James Y. Zhang and Xiaoming Shi and Pin-Yu Chen and Yuxuan Liang and Yuan-Fang Li and Shirui Pan and Qingsong Wen},
booktitle={The 12th International Conference on Learning Representations},
year={2024}
}

@article{gruver2024large,
  title={Large language models are zero-shot time series forecasters},
  author={Gruver, Nate and Finzi, Marc and Qiu, Shikai and Wilson, Andrew G},
  journal={Advances in Neural Information Processing Systems},
  volume={36},
  year={2024}
}

@article{rasul2023lag,
  title={Lag-llama: Towards foundation models for time series forecasting},
  author={Rasul, Kashif and Ashok, Arjun and Williams, Andrew Robert and Khorasani, Arian and Adamopoulos, George and Bhagwatkar, Rishika and Bilo{\v{s}}, Marin and Ghonia, Hena and Hassen, Nadhir Vincent and Schneider, Anderson and others},
  journal={arXiv preprint arXiv:2310.08278},
  year={2023}
}

@article{wang2024timemixer,
  title={{TimeMixer}: Decomposable multiscale mixing for time series forecasting},
  author={Wang, Shiyu and Wu, Haixu and Shi, Xiaoming and Hu, Tengge and Luo, Huakun and Ma, Lintao and Zhang, James Y and Zhou, Jun},
  journal={arXiv preprint arXiv:2405.14616},
  year={2024}
}

@misc{ekambaram2024ttms,
      title={{Tiny Time Mixers (TTMs)}: Fast Pre-trained Models for Enhanced Zero/Few-Shot Forecasting of Multivariate Time Series}, 
      author={Vijay Ekambaram and Arindam Jati and Pankaj Dayama and Sumanta Mukherjee and Nam H. Nguyen and Wesley M. Gifford and Chandra Reddy and Jayant Kalagnanam},
      year={2024},
      eprint={2401.03955},
      archivePrefix={arXiv},
      primaryClass={cs.LG}
}

@article{mirchandani2023large,
  title={Large language models as general pattern machines},
  author={Mirchandani, Suvir and Xia, Fei and Florence, Pete and Ichter, Brian and Driess, Danny and Arenas, Montserrat Gonzalez and Rao, Kanishka and Sadigh, Dorsa and Zeng, Andy},
  journal={arXiv preprint arXiv:2307.04721},
  year={2023}
}

@article{spathis2024first,
  title={The first step is the hardest: Pitfalls of representing and tokenizing temporal data for large language models},
  author={Spathis, Dimitris and Kawsar, Fahim},
  journal={Journal of the American Medical Informatics Association},
  volume={31},
  number={9},
  pages={2151--2158},
  year={2024},
  publisher={Oxford University Press}
}

@article{dettmers2024qlora,
  title={{QLoRA: Efficient Finetuning of Quantized LLMs}},
  author={Dettmers, Tim and Pagnoni, Artidoro and Holtzman, Ari and Zettlemoyer, Luke},
  journal={Advances in Neural Information Processing Systems},
  volume={36},
  year={2024}
}

@article{kang2024ina,
  title={{InA}: Inhibition Adaption on pre-trained language models},
  author={Kang, Cheng and Prokop, Jindrich and Tong, Lei and Zhou, Huiyu and Hu, Yong and Novak, Daniel},
  journal={Neural Networks},
  pages={106410},
  year={2024},
  publisher={Elsevier}
}

@article{zhou2021informer, title={Informer: Beyond Efficient Transformer for Long Sequence Time-Series Forecasting}, volume={35},  number={12}, journal={Proceedings of the AAAI Conference on Artificial Intelligence}, author={Zhou, Haoyi and Zhang, Shanghang and Peng, Jieqi and Zhang, Shuai and Li, Jianxin and Xiong, Hui and Zhang, Wancai}, year={2021}, pages={11106--11115} }

@article{tan2021time,
  title={Time series extrinsic regression: Predicting numeric values from time series data},
  author={Tan, Chang Wei and Bergmeir, Christoph and Petitjean, Fran{\c{c}}ois and Webb, Geoffrey I},
  journal={Data Mining and Knowledge Discovery},
  volume={35},
  number={3},
  pages={1032--1060},
  year={2021},
  publisher={Springer}
}

@article{jiang2023mistral,
  title={Mistral 7B},
  author={Jiang, Albert Q and Sablayrolles, Alexandre and Mensch, Arthur and Bamford, Chris and Chaplot, Devendra Singh and Casas, Diego de las and Bressand, Florian and Lengyel, Gianna and Lample, Guillaume and Saulnier, Lucile and others},
  journal={arXiv preprint arXiv:2310.06825},
  year={2023}
}

@article{touvron2023llama,
  title={{LLaMA}: Open and efficient foundation language models},
  author={Touvron, Hugo and Lavril, Thibaut and Izacard, Gautier and Martinet, Xavier and Lachaux, Marie-Anne and Lacroix, Timoth{\'e}e and Rozi{\`e}re, Baptiste and Goyal, Naman and Hambro, Eric and Azhar, Faisal and others},
  journal={arXiv preprint arXiv:2302.13971},
  year={2023}
}

@article{liu2019roberta,
  title={{RoBERTa}: A robustly optimized {BERT} pretraining approach},
  author={Liu, Yinhan and Ott, Myle and Goyal, Naman and Du, Jingfei and Joshi, Mandar and Chen, Danqi and Levy, Omer and Lewis, Mike and Zettlemoyer, Luke and Stoyanov, Veselin},
  journal={arXiv preprint arXiv:1907.11692},
  year={2019}
}

@inproceedings{liu2021ecg,
  title={{ECG}-based heart arrhythmia diagnosis through attentional convolutional neural networks},
  author={Liu, Ziyu and Zhang, Xiang},
  booktitle={2021 IEEE International Conference on Internet of Things and Intelligence Systems (IoTaIS)},
  pages={156--162},
  year={2021},
  organization={IEEE}
}

@article{seyfi2022generating,
  title={Generating multivariate time series with {CO}mmon {S}ource {C}oordInated {GAN} {(COSCI-GAN)}},
  author={Seyfi, Ali and Rajotte, Jean-Francois and Ng, Raymond},
  journal={Advances in Neural Information Processing Systems},
  volume={35},
  pages={32777--32788},
  year={2022}
}

@inproceedings{yang2021voice2series,
  title={Voice2series: Reprogramming acoustic models for time series classification},
  author={Yang, Chao-Han Huck and Tsai, Yun-Yun and Chen, Pin-Yu},
  booktitle={International conference on machine learning},
  pages={11808--11819},
  year={2021},
  organization={PMLR}
}

@article{EG19b,
  title   = {{ABBA: adaptive Brownian bridge-based symbolic aggregation of time series}},
  author  = {Elsworth, Steven and G\"{u}ttel, Stefan},
  year    = {2020},
  volume  = {34},
  pages   = {1175--1200},
  journal = {Data Mining and Knowledge Discovery}
}

@inproceedings{powers-1998-applications,
    title = "Applications and Explanations of {Z}ipf{'}s Law",
    author = "Powers, David M. W.",
    booktitle = "New Methods in Language Processing and Computational Natural Language Learning",
    year = "1998"
}

@inproceedings{vandenoord16_ssw,
  author={Aäron {van den Oord} and Sander Dieleman and Heiga Zen and Karen Simonyan and Oriol Vinyals and Alex Graves and Nal Kalchbrenner and Andrew Senior and Koray Kavukcuoglu},
  title={{WaveNet}: A Generative Model for Raw Audio},
  year=2016,
  booktitle={Proc. 9th ISCA Workshop on Speech Synthesis Workshop (SSW 9)},
  pages={125}
}

@misc{jin2024position,
      title={Position Paper: What Can Large Language Models Tell Us about Time Series Analysis}, 
      author={Ming Jin and Yifan Zhang and Wei Chen and Kexin Zhang and Yuxuan Liang and Bin Yang and Jindong Wang and Shirui Pan and Qingsong Wen},
      year={2024},
      eprint={2402.02713},
      archivePrefix={arXiv},
      primaryClass={cs.LG}
}

@misc{chen2024joint,
      title={Joint symbolic aggregate approximation of time series}, 
      author={Xinye Chen},
      year={2024},
      eprint={2401.00109},
      archivePrefix={arXiv},
      primaryClass={cs.DS}
}

@article{EG20b,
  title   = {{Time series forecasting using LSTM networks: 
              A symbolic approach}},
  author  = {Elsworth, Steven and G\"{u}ttel, Stefan},
  year    = {2020},
  pages   = {12},
  eprinttype    = {arXiv},
  eprint       = {2003.05672},
}

@InProceedings{10.1007/978-3-642-41398-8_24,
author="Malinowski, Simon
and Guyet, Thomas
and Quiniou, Ren{\'e}
and Tavenard, Romain",
title="{1d-SAX}: A Novel Symbolic Representation for Time Series",
booktitle="Advances in Intelligent Data Analysis XII",
year="2013",
}

@article{fABBA2022,
author = {Chen, Xinye and G\"{u}ttel, Stefan},
title = {An Efficient Aggregation Method for the Symbolic Representation of Temporal Data},
year = {2022},
publisher = {ACM},
journal = {ACM Transactions on Knowledge Discovery from Data}
}

@article{lin2007experiencing,
  author = {Lin, Jessica and Keogh, Eamonn and Wei, Li and Lonardi, Stefano},
  journal = {Data Mining and Knowledge Discovery},
  number = 2,
  pages = {107--144},
  publisher = {Springer},
  title = {Experiencing {SAX}: a novel symbolic representation of time series},
  volume = 15,
  year = 2007
}

@article{MAHAJAN201213,
title = {The planar k-means problem is NP-hard},
journal = {Theoretical Computer Science},
volume = {442},
pages = {13--21},
year = {2012},
note = {Special Issue on the Workshop on Algorithms and Computation (WALCOM 2009)},
author = {Meena Mahajan and Prajakta Nimbhorkar and Kasturi Varadarajan}
}

@inproceedings{10.1145/1374376.1374452,
author = {Dasgupta, Sanjoy and Freund, Yoav},
title = {Random Projection Trees and Low Dimensional Manifolds},
year = {2008},
publisher = {ACM},
booktitle = {Proceedings of the Fortieth Annual ACM Symposium on Theory of Computing},
pages = {537–546},
numpages = {10},
series = {STOC '08}
}

@ARTICLE{UCRArchive2018,
    author={Dau, Hoang Anh and Bagnall, Anthony and Kamgar, Kaveh and Yeh, Chin-Chia Michael and Zhu, Yan and Gharghabi, Shaghayegh and Ratanamahatana, Chotirat Ann and Keogh, Eamonn},
    journal={IEEE/CAA Journal of Automatica Sinica}, 
    title={The UCR time series archive}, 
    year={2019},
    volume={6},
    number={6},
    pages={1293--1305}
}

@ARTICLE{journals/tit/Lloyd82,
  author={Lloyd, S.},
  journal={IEEE Transactions on Information Theory}, 
  title={Least squares quantization in {PCM}}, 
  year={1982},
  volume={28},
  number={2},
  pages={129-137}
}

@article{Middlehurst2024,
  author    = {Middlehurst, Matthew and Sch{\"a}fer, Patrick and Bagnall, Anthony},
  title     = {Bake off redux: a review and experimental evaluation of recent time series classification algorithms},
  journal   = {Data Mining and Knowledge Discovery},
  year      = {2024},
  volume    = {38},
  pages     = {1958--2031},
  doi       = {10.1007/s10618-024-01022-1},
  url       = {https://doi.org/10.1007/s10618-024-01022-1}
}


\end{document}